\documentclass[numbers]{article}

\usepackage[eandd,final]{neurips_2026}

\usepackage[utf8]{inputenc}
\usepackage[T1]{fontenc}

\usepackage{hyperref}
\usepackage{url}
\usepackage{booktabs}
\usepackage{amsfonts}
\usepackage{amsmath}
\usepackage{nicefrac}
\usepackage{microtype}
\usepackage{xcolor}
\usepackage{graphicx}
\usepackage{enumitem}
\usepackage{xspace}
\usepackage{listings}
\usepackage{multirow}
\usepackage{colortbl}
\usepackage{natbib}
\usepackage{wrapfig}
\usepackage{float}
\hypersetup{hypertexnames=false,
  pdftitle={WorldCoder-Bench: Benchmarking Physically Grounded 3D World Synthesis},
  pdfauthor={Shuo Lu; Yinuo Xu; Kecheng Yu; Siru Jiang; Yongcan Yu; Yubin Wang; Haitao Yang; Yuxiang Zhang; Bin Wang; Ran He; Jian Liang}}

\newcommand{\bench}{\textsc{WorldCoder-Bench}\xspace}
\newcommand{\benchcore}{\textsc{WorldCoder-Core}\xspace}
\newcommand{\benchext}{\textsc{WorldCoder-Extended}\xspace}
\newcommand{\benchrob}{\textsc{WorldCoder-Robust}\xspace}
\newcommand{\benchdev}{\textsc{WorldCoder-Dev}\xspace}

\newcommand{\vcov}{\text{V-Cov}}
\newcommand{\acov}{\text{A-Cov}}
\newcommand{\scov}{\text{S-Cov}}
\newcommand{\tcov}{\text{T-Cov}}

\newcommand{\method}{\textsc{StateProbe}\xspace}
\definecolor{cRed}{RGB}{220,50,47}
\definecolor{cGreen}{RGB}{0,150,80}
\definecolor{cBlue}{RGB}{38,100,180}
\definecolor{cGray}{RGB}{120,120,120}
\definecolor{codebg}{RGB}{248,248,248}

\title{WorldCoder-Bench: Benchmarking Physically Grounded 3D World Synthesis}

\author{%
  \textbf{Shuo Lu}$^{1,\dagger}$\quad
  \textbf{Yinuo Xu}$^{1,\dagger}$\quad
  \textbf{Kecheng Yu}$^{1,2}$\quad
  \textbf{Siru Jiang}$^{1}$\quad
  \textbf{Yongcan Yu}$^{1}$\quad
  \textbf{Yubin Wang}$^{3}$\\
  \textbf{Haitao Yang}$^{3}$\quad
  \textbf{Yuxiang Zhang}$^{3}$\quad
  \textbf{Bin Wang}$^{3}$\quad
  \textbf{Ran He}$^{1,2}$\quad
  \textbf{Jian Liang}$^{1,2,\ddagger}$\\[6pt]
  \normalfont $^{1}$NLPR \& MAIS, Institute of Automation, Chinese Academy of Sciences (CASIA)\\
  \normalfont $^{2}$School of Artificial Intelligence, UCAS\hspace{0.8em}%
  \normalfont $^{3}$Huawei Noah's Ark Lab\\[0.1cm]
  \begingroup\hypersetup{hidelinks}%
  \normalfont\normalsize
  \href{https://github.com/shuolucs/WorldCoder-Bench}{\raisebox{-0.15em}{\includegraphics[height=0.95em]{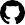}}\;\texttt{WorldCoder-Bench}}\quad
  \href{https://huggingface.co/datasets/shuolucs/WorldCoder-Bench}{\raisebox{-0.15em}{\includegraphics[height=0.95em]{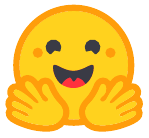}}\;\texttt{Dataset}}\quad
  \texttt{\{shuolucs, liangjian92\}@gmail.com}\endgroup\\[2pt]
}

\begin{document}
\maketitle

\renewcommand{\thefootnote}{\fnsymbol{footnote}}
\footnotetext[0]{
$^{\dagger}$Equal contribution.\quad 
$^{\ddagger}$Corresponding author.
}
\renewcommand{\thefootnote}{\arabic{footnote}}

\begin{abstract}
Large language models (LLMs) are increasingly asked not only to write static interfaces, but to construct executable interactive worlds from natural language. Browser-native 3D, commonly built with \texttt{Three.js}, is a natural next frontier: generated programs must integrate assets, obey spatial and physical constraints, and keep user-facing controls synchronized with hidden runtime state. Existing web-generation benchmarks and evaluators, however, largely observe only pixels or DOM nodes, while the mechanics of a \texttt{Three.js} world unfold inside an opaque \texttt{<canvas>}. 
We introduce \bench, a benchmark for autonomous, physically grounded 3D world synthesis. \bench contains $2{,}026$ expert-curated tasks across Simulation, Rendering, and Application scenarios, with optional \texttt{.glb} assets and hidden behavioral contracts. 
We further propose \method, an execution-based protocol that probes generated programs in a controlled browser and verifies hidden, mutation-hardened contracts over runtime states and transitions.
Beyond verification coverage, we report Return on Automation and Time Efficiency Multiplier to estimate quality-adjusted cost and time utility.
The best verification coverage is $27.8\%$ across nine models on \benchcore and $19.9\%$ across five on \benchrob, with failures dominated by state-schema drift and broken interaction chains rather than missing scene elements.
Utility metrics further show that cheap or fast models can still provide substantial value on easier domains. 
\end{abstract}

\section{Introduction}
\label{sec:intro}
Frontier large language models (LLMs) have moved past producing static UI mockups to generating executable, end-to-end web applications from a single instruction~\citep{design2code2025,webarena2023,artifactsbench2025}. A natural next step is browser-native 3D worlds: physics simulators, scientific visualizations, configurable products, and casual games shipped as a single HTML page~\citep{steenbergen2010analysis,dionisio20133d,sajja2025hydro3djs}. \texttt{Three.js}, with its deep WebGL integration and vast public corpus, has emerged as the dominant substrate for this kind of generation. Authoring such a world by hand is a multi-day frontend job; making an LLM author one in seconds promises a step change in how interactive 3D content is built. The moment the deliverable becomes a \emph{world} rather than a page, however, ``looks right'' stops being a useful proxy for ``works right'': objects must collide where they are supposed to, energy must dissipate at the right rate, and on-screen widgets must stay in lockstep with the engine state as the user interacts.

\begin{figure}[t]
  \centering
  \includegraphics[width=\linewidth]{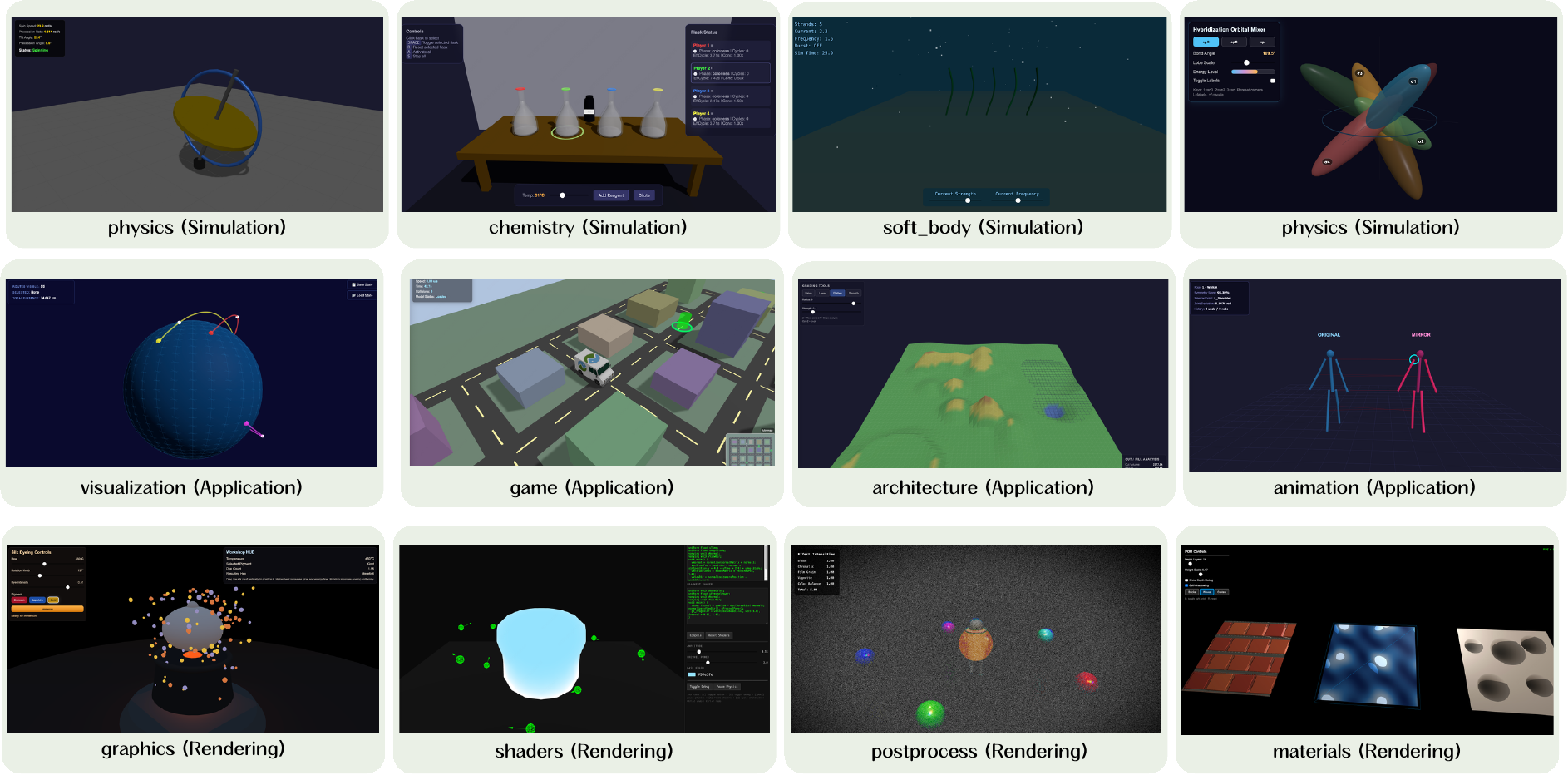}
  \caption{Representative generated worlds across Simulation, Application, and Rendering. Tasks combine visual presentation with physical or interactive requirements.}
  \label{fig:task_example}
\end{figure}

Whether the resulting code actually works is difficult to tell from the outside~\citep{fawzy2025vibe,feathers2004working}. \texttt{Three.js} renders into a \texttt{<canvas>}; the geometry, physics, animation phase, and interaction logic that determine correctness live inside the JavaScript runtime and are not directly exposed by screenshots or DOM structure. In our experiments, DOM-based scoring is essentially uncorrelated with contract-based correctness (per-pair Kendall $\tau_b{=}{-}0.02$ across $1{,}434$ pairs), and an 8-step agent inspecting the source at $\sim\!400\times$ the cost of DOM grants passing marks to $45.6\%$ of outputs with reference V-Cov below $30\%$. These results motivate direct runtime verification alongside surface-level evaluation.

We introduce \bench, a benchmark whose unit of evaluation is a fully executable \texttt{Three.js} world; representative examples are shown in Figure~\ref{fig:task_example}. \bench contains $2{,}026$ expert-curated tasks that span the major intents of browser-native 3D generation, from physically evolving simulations and controllable rendering effects to interactive goal-driven applications. These tasks cover $15$ fine-grained domains and include both primitive-only scenes and asset-dependent worlds with \texttt{.glb} resources. Each instance is provided as a structured directory with a natural-language brief, required interface schema, and optional assets, and the model must produce a single HTML page that loads, renders, and responds to user actions in a standard browser.

To pierce the canvas and judge what the world actually does, we pair the benchmark with \method, our execution-based state verification protocol. 
\method runs each generated program in a headless Chromium instance and reads a small runtime interface that exposes task-relevant variables. A scripted action sequence drives the world while \method snapshots state before and after each step and checks the resulting deltas against a hidden behavioral contract authored by domain experts. Critically, \method scrutinizes the contracts themselves through mutation testing: programmatically injected defects (deleted state updates, scaled physical constants, swapped event targets) test checker sensitivity to these failure modes. We complement the primary verification coverage metric with two utility multipliers, return on automation and time efficiency multiplier, which compare generation cost and time with estimated 3D-web developer labor.

\noindent\textbf{Summary of contributions.}
1) We present \bench, a benchmark that scores models on whether a generated \texttt{Three.js} world satisfies behavioral contracts rather than merely appearing visually plausible, comprising $2{,}026$ expert-curated tasks across three macro-categories and $15$ fine-grained domains.
2) We propose \method, an execution-based state-verification protocol driven by hidden behavioral contracts, with mutation testing to calibrate sensitivity to injected defects.
3) We introduce Return on Automation and Time Efficiency Multiplier, task-level utility estimates for 3D web development that combine inference cost and latency with public developer labor rates and discount low-coverage outputs.
4) We benchmark nine frontier models with \method: none exceeds $30\%$ Verification Coverage on Core, while an inspection agent grants passing marks to $45.6\%$ of low-reference-coverage outputs, motivating complementary state-level checks.

\section{\bench}
\label{sec:benchmark}

\begin{figure}[t]
  \centering
  \includegraphics[width=\linewidth]{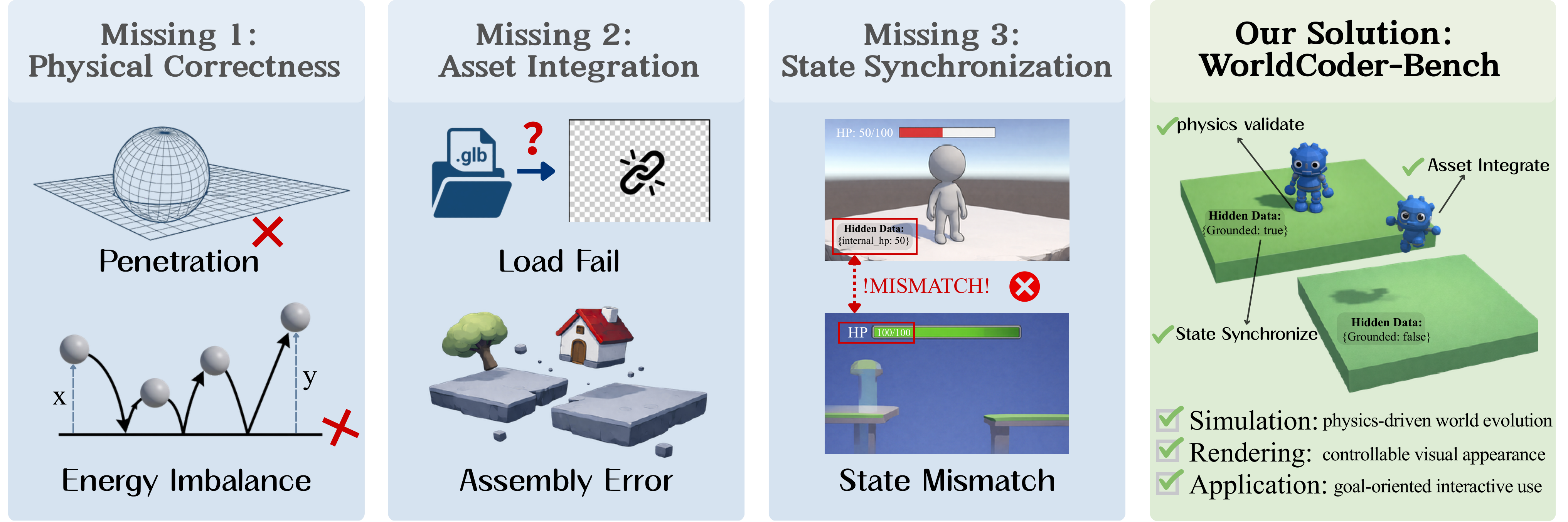}
  \caption{WorldCoder-Bench specifies executable tasks; StateProbe evaluates runtime behavior. Physical, asset, and synchronization requirements need complementary state and rendering evidence.}
  \label{fig:bench}
\end{figure}

\bench evaluates autonomous, physically grounded 3D world synthesis by
requiring models to generate self-contained \texttt{Three.js} programs from
natural-language specifications. As illustrated in Figure~\ref{fig:bench},
prior benchmarks largely overlook three aspects that are
indispensable for executable worlds: physical correctness~\citep{design2code2025,websight2024,frontendbench2025,webbench2025,artifactsbench2025,iwrbench2025,gamedevbench2026,vgamegym2025}, asset integration~\citep{design2code2025,websight2024,pix2code2018,interaction2code2025}, and state synchronization~\citep{design2code2025,websight2024,artifactsbench2025,frontendbench2025,iwrbench2025}. \bench targets all three
via \method, an execution-based state-verification protocol that scores generated worlds against hidden behavioral contracts. We
define the task, curation process, and dataset composition in this section;
\method is elaborated in Section~\ref{sec:eval}.

\subsection{Task Formulation}
\label{sec:task_def}

We formulate \bench as an end-to-end conditional code generation task for executable 3D world synthesis. Each task instance is a structured directory $x=(\mathcal{I}, \mathcal{A})$, where $\mathcal{I}$ is a \texttt{task.json} file containing the natural-language instruction, and $\mathcal{A}$ is an optional \texttt{assets/} folder containing pre-authored 3D resources such as \texttt{.glb} models. Given $x$, the model must output a single self-contained HTML file $y$ that uses \texttt{Three.js} to render and control a functional, interactive 3D world in a standard browser.
The generated program is executed in a browser and evaluated against held-out programmatic assertions. Generation requests include the visible task input, but omit evaluation contracts, action sequences, thresholds, and test scripts. This separates generation from evaluation while allowing correctness to be judged through executable runtime behavior rather than static appearance.

\subsection{Data Curation}
\label{sec:construction}

\begin{figure}[t]
  \centering
  \includegraphics[width=\linewidth]{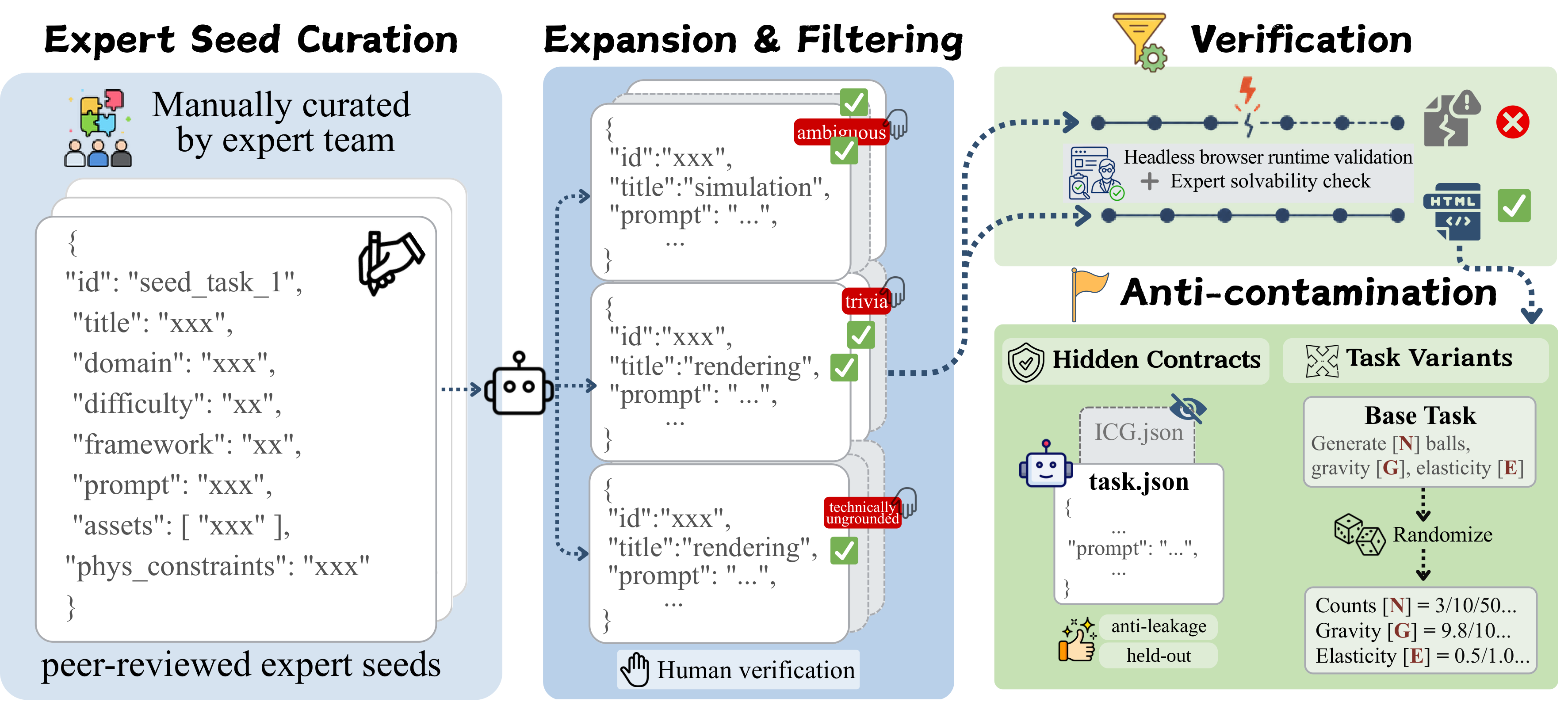}
  \caption{Task construction through expert seeds, LLM-assisted expansion and filtering, execution validation, contracts, and controlled variants.}
  \label{fig:pipeline}
\end{figure}

To ensure high quality, diversity, and rigorous evaluability, \bench is constructed through a systematic four-stage pipeline, as illustrated in Figure~\ref{fig:pipeline}.

\textbf{Stage I: Expert Seed Curation.} 
A team of five PhD researchers in 3D graphics and interactive systems spent two months hand-crafting seed tasks. Each seed undergoes strict peer review to guarantee objective clarity, implementation feasibility, and unambiguous interaction flows. These seeds serve as foundational templates for diverse task families rather than a separate evaluation split.

\textbf{Stage II: Expansion \& Filtering.} 
Using the expert seeds, we employ LLM-assisted expansion to generate over 10{,}000 candidate tasks covering various spatial layouts, physical mechanics, and rendering effects. Annotators then rigorously filter this pool, discarding tasks that are ambiguous, trivial, underspecified, or incapable of being operationalized into deterministic runtime checks.

\textbf{Stage III: Verification.} 
Surviving candidates undergo runtime validation in a headless browser to confirm asset availability, stable loading, and compatibility with our standardized interface. Experts then author behavioral evaluation contracts specifying expected affordances, reachable states, action-driven transitions, and state-level assertions. Tasks exhibiting unstable execution or unreliable evaluability are discarded (see Section~\ref{sec:eval}).
\begin{wrapfigure}{r}{0.45\textwidth}
  \centering  \includegraphics[width=\linewidth]{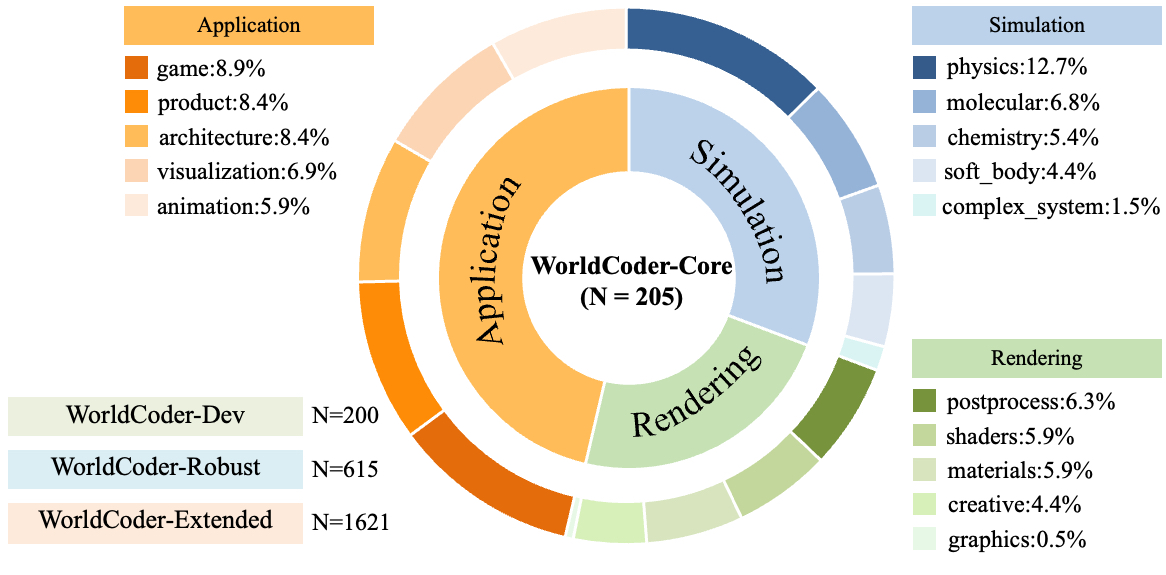}
  \caption{Distribution and composition of \bench.}
  \label{fig:benchmark_distribution}
\end{wrapfigure}

\textbf{Stage IV: Held-out Checks and Variants.}
The pipeline yields 2{,}026 finalized tasks. To reduce direct exposure to evaluation logic, contracts and assertions are excluded from generation requests; Dev contracts are released for debugging. Furthermore, we generate robustness variants by perturbing physical constants, object counts, initial states, and asset choices. This targeted randomization tests sensitivity to changes in task specifications rather than assuming that success on a fixed template establishes generalization.

\paragraph{Held-out behavioral ground truth.}
Correctness in \bench is behavioral rather than tied to a single reference HTML: multiple \texttt{Three.js} implementations may be valid if their runtime states satisfy the intended contract. Expert-authored rubrics and reference traces define the behavioral reference. Test contracts are held out at inference time, while \benchdev releases them for debugging and evaluator integration. Release-version visibility is discussed in Appendix~\ref{app:release}.

\subsection{Dataset Composition and Splits}
\label{sec:categories}

Figure~\ref{fig:benchmark_distribution} summarizes the taxonomy and split design of \bench. We organize tasks by the primary user intent of the generated 3D world---observing system evolution, controlling visual presentation, or completing an interactive goal---which correspond to three macro-categories: \emph{Simulation}, \emph{Rendering}, and \emph{Application}. This taxonomy is intuitive for users and aligned with our evaluator, since each intent requires different runtime contracts over dynamics, rendering states, or interaction logic.

After curation, the 2{,}026 canonical tasks are partitioned by evaluation purpose rather than construction source. \benchcore contains 205 tasks with held-out contracts and serves as the primary leaderboard split, selected for difficulty using reference-model T-Cov. \benchext contains 1{,}621 tasks with held-out contracts for broader model comparison. \benchrob is a perturbation-augmented stress-test built on top of \benchcore: each Core task is instantiated as three variants under controlled randomization of physical constants (e.g., gravity, elasticity, masses), object counts, initial states, prompt phrasing, and \texttt{.glb} asset filenames, yielding 615 variants in total. Contracts are adapted to the corresponding variants to test behavior under changed specifications; this does not by itself distinguish mechanism understanding from template adaptation. \benchdev contains 200 public tasks with released contracts and reference outputs for local debugging and evaluator integration.

Each task is additionally annotated with a difficulty level (D1--D6) and an asset tier. Difficulty ranges from basic scene initialization to multi-step logic, strict physical constraints, and state synchronization. Asset tiers distinguish primitive-only tasks, single-\texttt{.glb} tasks, and multi-asset tasks. Full domain counts, split statistics, difficulty distributions, asset tiers, prompt lengths, and estimated human effort are reported in Appendix~\ref{app:stats}.

\section{Evaluation}
\label{sec:eval}

In this section, we propose \method, the execution-based state-verification protocol that \bench uses to evaluate generated 3D worlds. We first explain why external observation is insufficient, then introduce mutation-hardened behavioral contracts as hidden ground truth, describe the execution protocol that probes runtime state, and finally define the capability and practical-utility metrics that \method reports.

\subsection{From External Observation to State Verification}
\label{sec:limitations}

\begin{figure}[t]
  \centering
  \includegraphics[width=\linewidth]{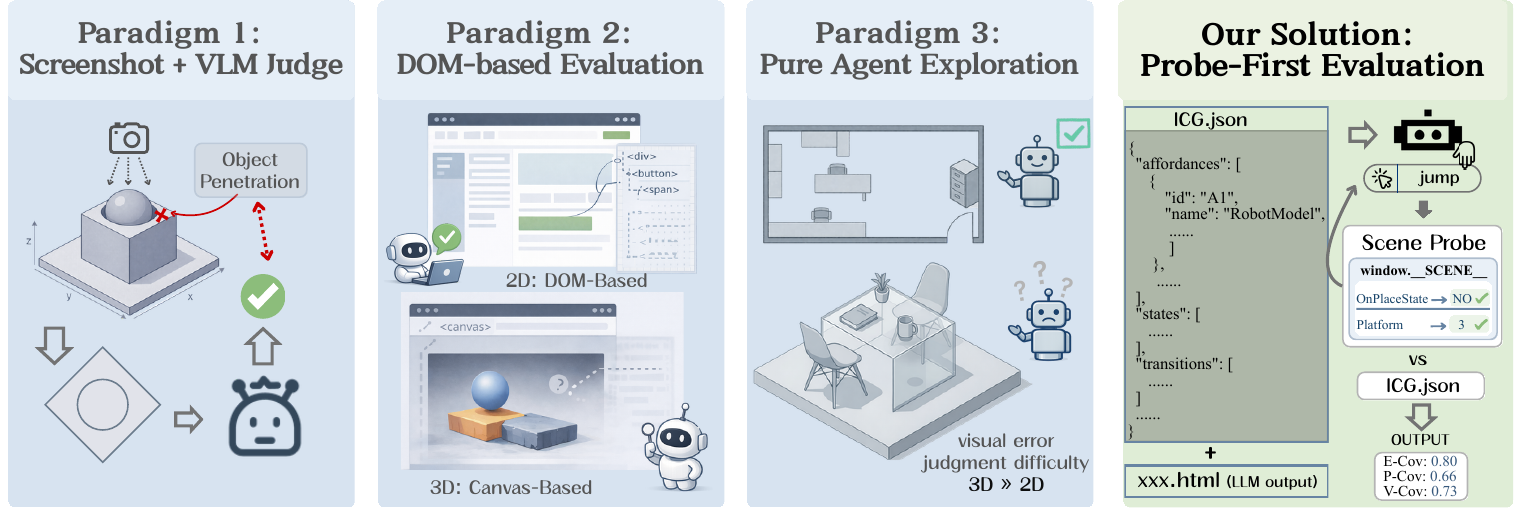}
  \caption{
    Evaluation paradigms for 3D world synthesis. External evaluators inspect pixels, DOM structure, or sampled traces and source; \method verifies hidden contracts through state probes.
  }
  \label{fig:eval}
\end{figure}

The core challenge is that correctness in a generated 3D world often lies beneath the rendered surface. As shown in Figure~\ref{fig:eval}, screenshot~\citep{design2code2025,pix2code2018,websight2024} or VLM judges~\citep{artifactsbench2025,webdevjudge2025} can assess visual plausibility~\citep{iwrbench2025,design2code2025}, but cannot reliably recover exact coordinates, velocities, collision states, conservation quantities, or hidden interaction variables. DOM-based evaluation works for ordinary web pages, but \texttt{Three.js} worlds are rendered as WebGL pixels inside \texttt{<canvas>}; the DOM exposes only peripheral elements such as buttons or labels. Agent exploration adds interaction and source inspection, but remains costly and dependent on sampled evidence.
These paradigms do not enforce the task's runtime contract. \method instead makes runtime behavior observable: it reads internal state variables, applies controlled actions, and checks whether the world satisfies the physical, spatial, and interactive conditions specified by the task.

\subsection{Mutation-Hardened Contracts}
\label{sec:contracts}

\textbf{Expert-Authored Behavioral Contracts.} 
Each task is paired with a hidden executable specification that defines the required affordances, states, transitions, and assertions. Authored by domain experts based on the task prompt, these contracts encode expected objects and controls, canonical interaction paths, numerical tolerances, and task-specific invariants (e.g., physical conservation laws, rendering-state changes, or UI-state synchronization) that a correct world must satisfy.

\textbf{Calibration via Mutation Testing.} 
To reduce permissive evaluation, \method calibrates contracts using mutation testing inspired by software engineering practices~\citep{jia2011mutation}. Starting from internal reference implementations, we deliberately inject common failures into the HTML---such as stale state updates, corrupted physical constants, broken asset loading, swapped event targets, or HUD mismatches. Surviving mutants identify checks for revision or tasks for exclusion. This hardening step tests sensitivity to injected defects; a \textsc{Check\_Pass} certifies the scheduled checks, not all possible behaviors.

\subsection{Execution Protocol}
\label{sec:protocol}

\textbf{Runtime state interface.}
Each generated program must expose a standardized runtime state interface, typically a global object such as \texttt{window.\_\_3D\_STATE\_\_}. This interface reports raw variables such as object coordinates, velocities, selected modes, counters, camera distances, and interaction flags. Generation requests include the interface schema but omit the contract's action sequence, thresholds, expected transitions, and assertion scripts.

\textbf{Browser execution.}
\method runs each generated program in a headless Chromium browser managed by Playwright, with fixed browser settings and a local version-locked \texttt{Three.js} archive. \method first checks executability: the page must load, create a live WebGL context, maintain a render loop, and avoid early JavaScript exceptions. Failure at this stage yields \textsc{Runtime\_Crash}.

\textbf{Action-driven probes.}
For executable programs, \method applies deterministic action-driven probes. For each scripted action, such as a click, key press, drag event, or physics-step command, \method records a \emph{before} snapshot, applies the action, records an \emph{after} snapshot, and checks hidden assertions over the resulting state delta. Each task is then labeled \textsc{Runtime\_Crash}, \textsc{Check\_Fail}, or \textsc{Check\_Pass}, separating load-time failures, behavioral violations, and executions passing scheduled checks while retaining an auditable trace of every probed transition.

\subsection{Metrics}
\label{sec:metrics}

A binary outcome is too coarse for complex 3D worlds, so \method reports four coverage metrics aligned with the contract structure:
\begin{itemize}[leftmargin=*,itemsep=2pt,parsep=0pt]
    \item \emph{Affordance Coverage} ($\acov$): fraction of specified affordance checks passed.
    \item \emph{State Coverage} ($\scov$): fraction of specified state checks passed.
    \item \emph{Transition Coverage} ($\tcov$): fraction of transitions passing their postcondition checks.
    \item \emph{Verification Coverage} ($\vcov$): the tier-weighted transition-verification score.
\end{itemize}
$\vcov$ is the primary leaderboard metric (score policy in Appendix~\ref{app:metric_policy}); $\acov$, $\scov$, and $\tcov$ diagnose where a generated world fails. We also report crash rate, missing-output rate, token usage, cost, and latency.

\paragraph{Quality-adjusted utility.}
To estimate practical value, we combine correctness, cost, and time. For each task $t$, the normalized verification score $\widehat{\vcov}(t)\in[0,1]$ acts as a quality discount. Let $H_{\text{human}}(t)$ be the expert-estimated human completion time, $R$ the 3D-web developer hourly rate,\footnote{
We set $R=\$60$/hour based on public salary estimates for WebGL and web developers: Wellfound reports an average WebGL developer salary of roughly $\$125{,}000$/year, or $\$60.10$/hour under a 2,080-hour work year~\citep{wellfound_webgl_salary}; Talent.com reports an average U.S. web developer wage of $\$61.83$/hour~\citep{talent_web_developer_salary}.
} $C_{\text{model}}(t)$ the model cost, and $H_{\text{model}}(t)$ the model generation time:
\begin{align}
  \text{RoA} &=
  \frac{\sum_{t=1}^{n} \widehat{\vcov}(t)\cdot H_{\text{human}}(t)\cdot R}
       {\sum_{t=1}^{n} C_{\text{model}}(t)},
  &
  \text{TEM} &=
  \frac{\sum_{t=1}^{n} \widehat{\vcov}(t)\cdot H_{\text{human}}(t)}
       {\sum_{t=1}^{n} H_{\text{model}}(t)}.
\end{align}
Return on Automation ($\text{RoA}$) reports estimated quality-adjusted human-labor value per dollar of model cost; Time Efficiency Multiplier ($\text{TEM}$) reports estimated quality-adjusted human-hours per generation hour. Both discount outputs by contract coverage; neither measures realized labor savings.

\begin{table*}[!t]
\centering
\definecolor{firstbg}{HTML}{FFCCCC}
\definecolor{secondbg}{HTML}{CCE5FF}

\caption{Core results (205 planned tasks). Coverage is in percent; V-Cov is the archived transition-weighted score. RoA and TEM are secondary proxies in \$/\$ and hr/hr. Colors mark the best/second-best in each column.}
\label{tab:main_results}
\vspace{5pt}
\renewcommand{\arraystretch}{1.2}
\resizebox{\textwidth}{!}{%
\begin{tabular}{l|c|ccc|ccc|cc}
\hline
\multirow{2}{*}{\textbf{Model}} & \multirow{2}{*}{\textbf{V-Cov (\%) $\uparrow$}} & \multicolumn{3}{c|}{\textbf{V-Cov by Macro-Category (\%)}} & \multicolumn{3}{c|}{\textbf{Diagnostic Coverages (\%)}} & \multicolumn{2}{c}{\textbf{Utility proxies}} \\
\cline{3-10}
& & \textbf{Sim.} & \textbf{Render.} & \textbf{App.} & \textbf{A-Cov} & \textbf{S-Cov} & \textbf{T-Cov} & \textbf{RoA $\uparrow$} & \textbf{TEM $\uparrow$} \\
\hline
\multicolumn{10}{c}{\textit{Proprietary Models}} \\
\hline
GPT-5.4~\citep{gpt5.4-2026}                       & \cellcolor{firstbg}\textbf{27.8} & \cellcolor{secondbg}26.8 & \cellcolor{firstbg}\textbf{37.8} & \cellcolor{firstbg}\textbf{23.5} & \cellcolor{secondbg}66.2 & \cellcolor{firstbg}\textbf{46.1} & \cellcolor{firstbg}\textbf{26.3} & \phantom{0}1{,}264.6 & \cellcolor{firstbg}\textbf{68.87} \\
Gemini 3.1 Pro Preview~\citep{gemini2026}         & \cellcolor{secondbg}26.5 & \cellcolor{firstbg}\textbf{27.0} & \cellcolor{secondbg}35.9 & 21.4 & \cellcolor{firstbg}\textbf{67.4} & \cellcolor{secondbg}45.9 & \cellcolor{secondbg}25.1 & \phantom{0}1{,}551.2 & \cellcolor{secondbg}56.81 \\
Claude Sonnet 4.6~\citep{claude-sonnet-4.6-2026}  & 18.7 & 21.4 & 25.1 & 13.7 & 56.5 & 33.4 & 17.0 & \phantom{0}1{,}020.7 & 48.34 \\
Claude Opus 4.6~\citep{claude-Opus-4.6-2026}      & 17.5 & 18.2 & 22.2 & 14.6 & 58.6 & 38.3 & 15.8 & \phantom{00,}562.3 & 44.19 \\
\hline
\multicolumn{10}{c}{\textit{Additional API-Served Models}} \\
\hline
Qwen3.6-Plus~\citep{qwen3.6-plus-2026}            & 25.3 & 24.6 & 30.4 & \cellcolor{secondbg}23.2 & 65.3 & 43.6 & 23.7 & \phantom{0}4{,}343.9 & 37.10 \\
DeepSeek-V3.2~\citep{deepseek-v3.2-2026}          & 21.9 & 21.3 & 28.8 & 18.8 & 60.5 & 37.4 & 20.7 & \cellcolor{secondbg}14{,}538.8 & 22.49 \\
DeepSeek-V4-Flash~\citep{deepseek-v4-flash-2026}  & 16.5 & 11.8 & 24.2 & 15.8 & 40.6 & 30.7 & 15.9 & \cellcolor{firstbg}\textbf{19{,}150.4} & 23.23 \\
Kimi-K2.5~\citep{kimi-k2.5-2026}                  & 18.2 & 13.6 & 29.3 & 15.8 & 50.9 & 34.4 & 17.1 & \phantom{0}4{,}062.7 & 38.26 \\
MiniMax-M2.7~\citep{minimax-m2.7-2026}            & 23.0 & 22.6 & 30.2 & 19.7 & 57.6 & 39.9 & 21.7 & 12{,}322.2 & 48.36 \\
\hline
\end{tabular}}
\end{table*}

\section{Experiments}
\label{sec:experiments}

We evaluate nine frontier models on \bench under a unified zero-shot protocol, characterizing their capability, robustness to parameter perturbations, economic value relative to expert labor, and dominant failure modes, four facets that together explain the gap between visible plausibility and behavioral correctness in LLM-generated 3D worlds.

\subsection{Setup}
\label{sec:exp_setup}

\textbf{Models.} We benchmark nine frontier models drawn from seven families: GPT-5.4~\citep{gpt5.4-2026}, Claude Opus 4.6 and Sonnet 4.6~\citep{claude-Opus-4.6-2026,claude-sonnet-4.6-2026}, Gemini 3.1 Pro Preview~\citep{gemini2026}, DeepSeek V3.2~\citep{deepseek-v3.2-2026} and V4-Flash~\citep{deepseek-v4-flash-2026}, Qwen 3.6-Plus~\citep{qwen3.6-plus-2026}, Kimi K2.5~\citep{kimi-k2.5-2026}, and MiniMax M2.7~\citep{minimax-m2.7-2026}. All nine appear in Table~\ref{tab:main_results}; system-level diagnostics (crash rate, missing-probe rate, mean generation tokens and API latency) are in Appendix~\ref{app:leaderboard}. Every model runs zero-shot on the same \texttt{task.json} and \texttt{assets} directory, emits a single self-contained HTML, and is scored by \method against the mutation-hardened contract of \S\ref{sec:contracts}. Models receive the task prompt, observable-state requirements, and required runtime interface schema, but not the contract checks, hidden action sequences, thresholds, or assertion scripts.

\begin{wraptable}{r}{0.7\textwidth}
\vspace{-10pt}
\centering
\small
\setlength{\tabcolsep}{3pt}
\renewcommand{\arraystretch}{1.1}
\caption{V-Cov (\%) on Core and Extended, with gains ($\Delta$, percentage points) and rank changes (Core $\to$ Extended).}
\label{tab:extended_results}
\resizebox{\linewidth}{!}{%
\begin{tabular}{lrrrr}
\toprule
\textbf{Model} & \textbf{Core} & \textbf{Extended} & \textbf{$\Delta$ (pp)} & \textbf{Rank} \\
\midrule
GPT-5.4~\citep{gpt5.4-2026}
    & \textbf{27.8} & \textbf{42.1} & +14.3 & $1\to1$ \\
Gemini 3.1 Pro Preview~\citep{gemini2026}
    & 26.5 & 40.6 & +14.1 & $2\to2$ \\
Qwen3.6-Plus~\citep{qwen3.6-plus-2026}
    & 25.3 & 37.8 & +12.5 & $3\to4$ \\
MiniMax-M2.7~\citep{minimax-m2.7-2026}
    & 23.0 & 38.9 & +15.9 & $4\to3$ \\
DeepSeek-V3.2~\citep{deepseek-v3.2-2026}
    & 21.9 & 35.4 & +13.5 & $5\to5$ \\
Claude Sonnet 4.6~\citep{claude-sonnet-4.6-2026}
    & 18.7 & 31.2 & +12.5 & $6\to8$ \\
Kimi-K2.5~\citep{kimi-k2.5-2026}
    & 18.2 & 29.6 & +11.4 & $7\to9$ \\
Claude Opus 4.6~\citep{claude-Opus-4.6-2026}
    & 17.5 & 33.8 & +16.3 & $8\to7$ \\
DeepSeek-V4-Flash~\citep{deepseek-v4-flash-2026}
    & 16.5 & 34.7 & +18.2 & $9\to6$ \\
\midrule
Nine-model mean
    & 21.7 & 36.0 & +14.3 & -- \\
\bottomrule
\end{tabular}%
}
\vspace{-8pt}
\end{wraptable}

\textbf{Metrics.} Capability is reported as Verification Coverage ($\vcov$, a tier-weighted transition-verification score) alongside Affordance, State, and Transition coverage (\S\ref{sec:metrics}). Practical utility is reported as Return on Automation (RoA, \$/\$) and Time Efficiency Multiplier (TEM, hr/hr), computed from logged tokens under the reported token-rate assumptions, measured API latency, and a difficulty-conditioned $H_{\text{human}}$ estimator anchored to seven expert annotations (median $4.7$\,h). These estimates are not measured developer completion times. Per-model rates, the fallback table, the estimated human-effort schedule, and an $R\!\in\!\{\$25,\$60,\$100\}$ sensitivity check are detailed in Appendix~\ref{app:human}. The Extended evaluations are detailed in Appendix~\ref{app:review_results}.

\subsection{Performance on \bench}
\label{sec:capability}

We evaluate nine frontier models on the deliberately difficult \benchcore split and the broader curated Extended split to assess their ability to generate functionally correct browser-native 3D worlds. Five patterns emerge.

\textbf{Functional correctness remains challenging on Core.}
GPT-5.4 leads with $\vcov{=}27.8\%$, followed by Gemini 3.1 Pro Preview ($26.5\%$), Qwen 3.6-Plus ($25.3\%$), and MiniMax M2.7 ($23.0\%$). The top four span only $4.8$ percentage points, with Qwen trailing GPT by $2.5$ points. No model exceeds $30\%$ on Core's held-out contracts. These results expose incomplete behavior on a deliberately difficult diagnostic split; they do not imply uniformly low coverage across all browser-native 3D requests.

\textbf{Affordance checks and behavioral checks expose different failure modes.}
The diagnostic columns distinguish observable affordances from state and transition correctness: $\acov$ ranges from $40.6\%$ to $67.4\%$, $\scov$ from $30.7\%$ to $46.1\%$, and $\tcov$ from $15.8\%$ to $26.3\%$. A control can exist without updating the world, and a HUD update can leave the underlying engine state stale. The lower transition coverage highlights the difficulty of implementing action-driven behavior. 

\textbf{Rendering, Simulation, and Application impose different demands.}
Rendering achieves higher V-Cov ($22.2$--$37.8\%$) than Simulation ($11.8$--$27.0\%$) and Application ($13.7$--$23.5\%$). Standard \texttt{Three.js} APIs, including \texttt{RawShaderMaterial} and \texttt{EffectComposer}, may facilitate material, shader, and post-processing tasks, whereas Simulation adds numerical correctness requirements and Application adds multi-step logic and tight state synchronization. Appendix~\ref{app:leaderboard} reports finer-grained domain variation.

\begin{table}[t]
\centering
\caption{Comparison archive ($1{,}434$ pairs). $\tau_b$ is correlation with its contract reference; LP is the share with reference score below $30\%$ called passing. Cost is relative to DOM-only. The reference row is definitional, not independent validation.}
\label{tab:paradigm_comparison}
\renewcommand{\arraystretch}{1.2}
\begin{tabular}{lcccc}
\toprule
\textbf{Paradigm} & \textbf{Pass rate (\%)} & \textbf{Kendall $\tau_b$ vs.\ $\vcov$} & \textbf{LP (\%) $\downarrow$} & \textbf{Cost} \\
\midrule
DOM-only inspection            & 32.6 & $-0.021$ & 33.7 & $1.0\times$ \\
Screenshot+VLM (GPT-5.4)       & \phantom{0}2.7  & $-0.068$ & \phantom{0}3.2  & ${\sim}50\times$ \\
Pure Agent & 49.7 & $+0.111$ & 45.6 & ${\sim}400\times$ \\
\midrule
\textbf{Contract reference} & --- & 1.000 & 0.0 & ${\sim}2\times$ \\
\bottomrule
\end{tabular}
\vspace{-13pt}
\end{table}

\textbf{Extended broadens coverage while preserving substantial rank agreement.}
All nine models achieve higher V-Cov on Extended (Table~\ref{tab:extended_results}). Mean V-Cov rises from $21.7\%$ on Core to $36.0\%$ on Extended, an increase of $14.3$ percentage points, with GPT-5.4 remaining first at $42.1\%$. Cross-split rankings are strongly correlated (Spearman $\rho{=}0.833$). This pattern is consistent with the difficulty-based split selection; both splits are curated rather than random samples of browser-native 3D requests.
\begin{wraptable}{r}{0.7\textwidth}
\vspace{-12pt}
\centering
\small
\caption{Robustness on \benchrob (\%).}
\label{tab:robustness}
\resizebox{\linewidth}{!}{%
\begin{tabular}{lcccc}
\toprule
\textbf{Model} & \textbf{V-Cov $\uparrow$} & \textbf{A-Cov} & \textbf{S-Cov} & \textbf{T-Cov} \\
\midrule
GPT-5.4~\citep{gpt5.4-2026}            & \textbf{19.9} & 65.6 & 44.9 & \textbf{18.3} \\
Gemini-3.1-Pro~\citep{gemini2026}     & 19.1 & \textbf{66.8} & \textbf{46.3} & 17.6 \\
Kimi-K2.5~\citep{kimi-k2.5-2026}          & 15.6 & 57.9 & 34.9 & 14.3 \\
Claude-Opus-4.6~\citep{claude-Opus-4.6-2026}    & 13.9 & 58.3 & 39.8 & 12.4 \\
Qwen3.6-Plus~\citep{qwen3.6-plus-2026}       & 13.5 & 52.2 & 34.8 & 12.1 \\
\bottomrule
\end{tabular}%
}
\vspace{-8pt}
\end{wraptable}

\textbf{External judgments capture different properties of generated worlds.}
We compare three external evaluation paradigms against the contract reference using $1{,}434$ $\langle\text{model},\text{task}\rangle$ pairs from seven models: $204$ pairs for Qwen and $205$ for each of the other six models (Table~\ref{tab:paradigm_comparison}). DOM-only yields Kendall $\tau_b{=}{-}0.021$, Screenshot$+$VLM yields $\tau_b{=}{-}0.068$, and an eight-step inspection agent yields $\tau_b{=}0.111$. The agent nevertheless labels $45.6\%$ of low-$\vcov$ outputs as passing. With DOM-only normalized to $1\times$ cost, Screenshot$+$VLM costs approximately $50\times$ and the agent approximately $400\times$; these ratios are not relative to \method. The weak rank agreement suggests that DOM presence, visual fidelity, and sampled interactions are imperfect proxies for contract-verified functional correctness. Appendix~\ref{app:paradigm_per_model} provides error analyses and details.
\subsection{Robustness on \benchrob}
\label{sec:robustness}

GPT-5.4 and Gemini lead Robust V-Cov at $19.9\%$ and $19.1\%$ (Table~\ref{tab:robustness}). GPT also leads T-Cov, while Gemini leads A-Cov and S-Cov, separating presence and reachability from completed transitions. Qwen drops from $25.3\%$ on Core to $13.5\%$ on Robust, indicating sensitivity to modified tasks.

Variants change prompts, physical parameters, counts, initial states, or assets while preserving the evaluation purpose. These five-model results measure sensitivity to modified specifications; they do not isolate memorization or template recall, and perturbation types vary by task.

\subsection{Practical Benefits}
\label{sec:practical}
\begin{wrapfigure}{r}{0.7\textwidth}
    \vspace{-10pt}
    \centering
    \includegraphics[width=\linewidth]{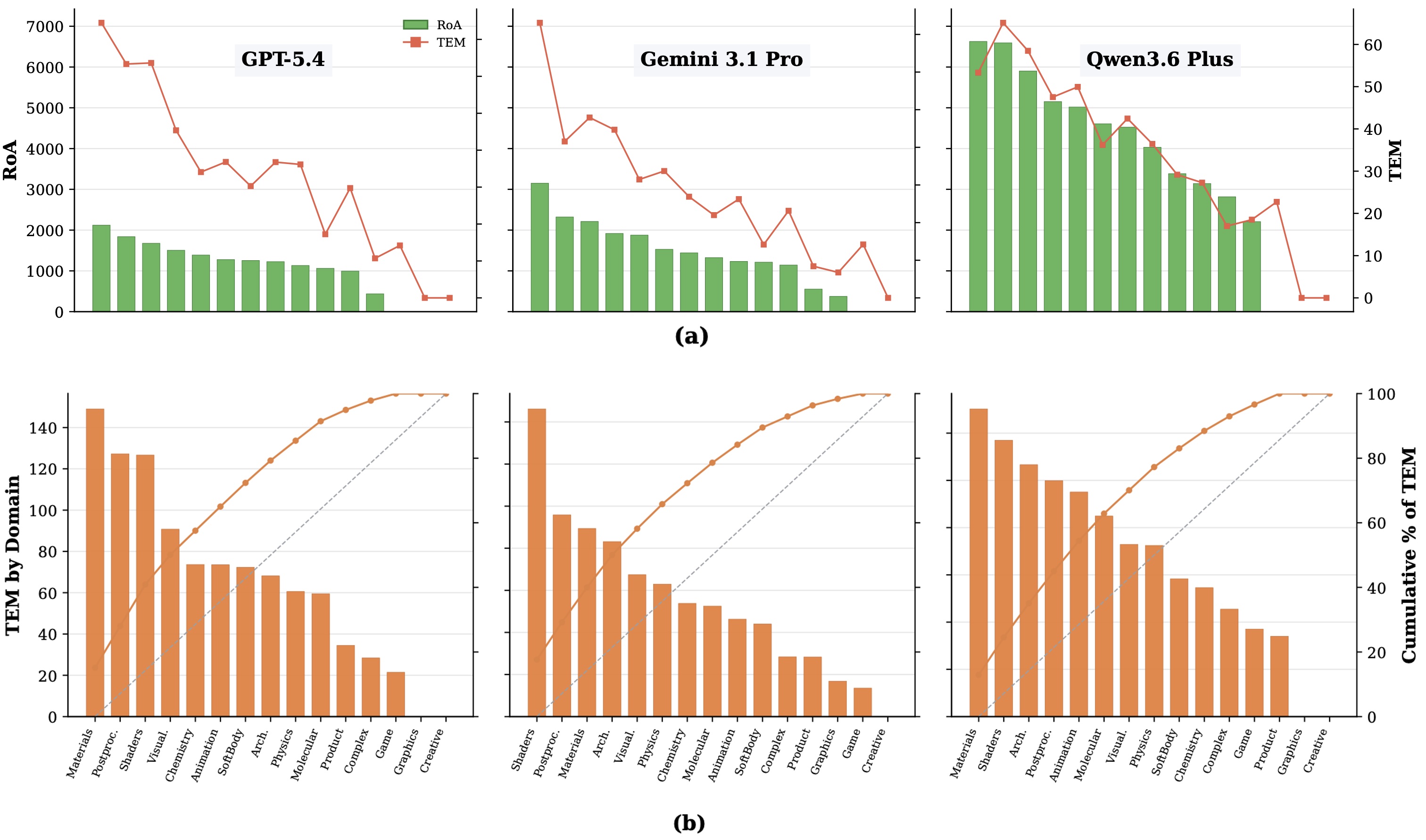}
    \vspace{-10pt}
    \caption{Per-domain RoA (a) and TEM Pareto curves (b) on \benchcore for three models.}
    \label{fig:roa_tem_pareto}
    \vspace{-15pt}
\end{wrapfigure}
Even at Core $\vcov$ below $30\%$, RoA and TEM (\S\ref{sec:metrics}) can be large under the assumed human-effort and token-rate schedule. These are quality-adjusted utility estimates.

\textbf{Aggregate utility flips the V-Cov ranking.}
The Economics columns of Table~\ref{tab:main_results} differ from the V-Cov order: DeepSeek-V4-Flash tops RoA at $\$19{,}150$/\$, about $12.3\!\times$ the strongest proprietary model (Gemini), with DeepSeek-V3.2 second ($\$14{,}539$/\$). GPT-5.4 leads TEM at $68.9\times$, combining high coverage with short generation latency; Gemini is second at $56.8\times$. These estimates favor DeepSeek on cost and GPT-5.4 or Gemini on quality-adjusted generation time.

\textbf{Per-domain decomposition exposes a Pareto profile.}
Figure~\ref{fig:roa_tem_pareto} decomposes GPT-5.4, Gemini-3.1-Pro, and Qwen3.6-Plus across the $15$ fine-grained domains; a seven-model profile is in Appendix~\ref{app:roa_tem_full}. (i) Materials, Shaders, and Postprocess contribute strongly to both metrics. (ii) Cumulative TEM bends above the uniform reference: the top-$5$ of $15$ domains contribute $55$--$65\%$ of total TEM. (iii) Complex System, Creative, and Graphics have low utility relative to these leading domains, highlighting a challenging tail.

\subsection{Error Analysis}
\label{sec:errors}

\begin{figure}[ht]
    \centering
    \includegraphics[width=0.95\linewidth]{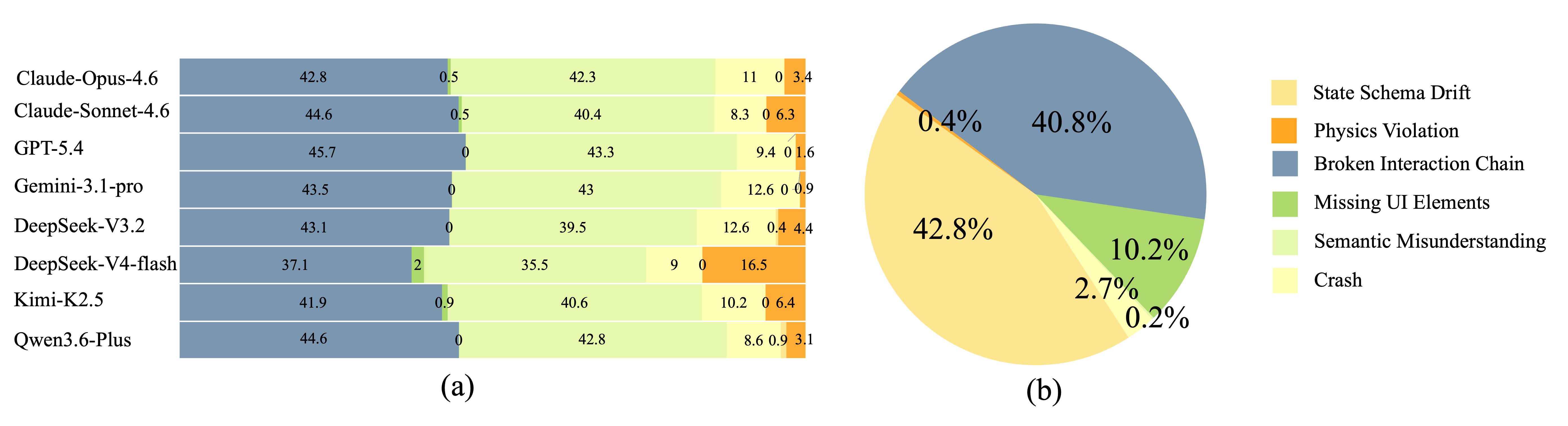}
    \caption{Error taxonomy and archived model diagnostic profiles. Primary-summary shares and non-exclusive record prevalence use different denominators; these statistics are not population failure-rate estimates.}
    \label{fig:error_analysis}
\end{figure}

To characterize dominant failure modes, we analyze $596$ zero-$\tcov$ records across nine models. The diagnostic taxonomy comprises five behavioral labels: \textbf{State Schema Drift} (deviation from prompt-specified field names or types), \textbf{Physics Violation} (failure of tested numerical invariants, such as energy conservation or stability), \textbf{Broken Interaction Chain} (user actions disconnected from exposed runtime state), \textbf{Missing UI Elements} (omission of requested controls), and \textbf{Semantic Misunderstanding} (incorrect task logic or generative algorithms). \textbf{Crash} is recorded separately as an execution signal. Labels may co-occur, so primary-summary shares and multi-label prevalence are not additive (Figure~\ref{fig:error_analysis}; Appendix~\ref{app:errors_full}).

\paragraph{Schema drift and broken interaction chains dominate the diagnostic summary.}
\textbf{State Schema Drift} ($42.8\%$) and \textbf{Broken Interaction Chain} ($40.8\%$) together account for $83.6\%$ of the primary-summary distribution. This pattern is consistent with the presence-versus-behavior gap in \S\ref{sec:capability}: models often produce visually plausible scenes and expose intermediate states, yet the required runtime interface remains incomplete, incorrectly structured, or unsynchronized with user actions. Complete omission of a state object differs from partial schema errors, while stale fields differ from an interface that updates only the HUD without changing the underlying engine. Interaction failures also include handlers bound to the wrong event and resets that leave related variables unchanged. These diagnoses distinguish observable failure modes without establishing their causal mechanisms.

\begin{table}[!htbp]
\centering
\small
\caption{Core V-Cov (\%) before and after interface completion. Omissions indicate absent probes.}
\label{tab:interface_completion}
\begin{tabular}{lrrrr}
\toprule
\textbf{Model} & \textbf{Omissions} & \textbf{Original} & \textbf{Completed} & \textbf{$\Delta$ (pp)} \\
\midrule
Kimi-K2.5 & 21/205 & 18.2 & 20.1 & +1.9 \\
Claude Opus 4.6 & 15/205 & 17.5 & 18.9 & +1.4 \\
\bottomrule
\end{tabular}
\end{table}

\paragraph{Completing omitted probes leaves substantial coverage gaps.}
The interface-completion ablation raises Core-average V-Cov from $18.2\%$ to $20.1\%$ for Kimi and from $17.5\%$ to $18.9\%$ for Opus (Table~\ref{tab:interface_completion}). These gains of $1.9$ and $1.4$ percentage points leave substantial coverage gaps. The ablation tests omitted-interface completion, not arbitrary program repair (Appendix~\ref{app:review_results}).

\paragraph{Crash incidence and interface correctness capture distinct failures.}
Using all $205$ planned tasks as the denominator, recorded crash rates are $3.4\%$ for GPT-5.4, $2.0\%$ for Gemini, and $42.4\%$ for DeepSeek-V4-Flash. These task-level rates are distinct from diagnostic shares within the zero-$\tcov$ archive. Low crash incidence indicates fewer runtime terminations, but does not imply a correct or action-synchronous state interface. Kimi-K2.5, for example, frequently omits the required state object entirely. Representative cases and their task and trace scopes are documented in Appendix~\ref{app:case_study}.

\paragraph{Physical and algorithmic reasoning remain a challenging tail.}
\textbf{Physics Violation} ($0.4\%$) and \textbf{Semantic Misunderstanding} ($2.7\%$) occupy smaller shares of the primary diagnostic summary. Their relative rarity in this zero-$\tcov$ subset should not be interpreted as evidence of reliable physical or algorithmic reasoning: Creative tasks expose incorrect generative logic, while Physics tasks remain sensitive to numerical precision and stability.


\section{Conclusion}
\label{sec:conclusion}

We introduced \bench, a benchmark for autonomous, physically grounded 3D world synthesis, and \method, an execution-based protocol that evaluates generated \texttt{Three.js} worlds against hidden, mutation-hardened behavioral contracts. The best Verification Coverage is $27.8\%$ across nine models on \benchcore and $19.9\%$ across five on \benchrob, while DOM, Screenshot+VLM, and agent-based evaluators show weak agreement with the contract-based ranking. The diagnostic summary is dominated by state-schema drift and broken interaction chains. RoA and TEM estimate quality-adjusted utility on easier domains under the stated cost and human-effort assumptions.
We hope \bench helps pave the way for autonomous systems that can construct executable worlds, not just plausible scenes.

\section{Acknowledgments}
This work was funded by the Beijing Natural Science Foundation Z260008, the National Cyber Security-National Science and Technology Major Project 2026ZD1500300, and the National Natural Science Foundation of China under Grants U2441251 and 62276256.

\clearpage
\bibliographystyle{worldcoder-compact}
\bibliography{neurips}

\newpage

\appendix

\section{Related Work}
\label{sec:related}

\subsection{Web and Game Code Generation}

Web benchmarks span screenshot reconstruction~\citep{design2code2025,websight2024,pix2code2018}, modern frontend tasks~\citep{webuibench2025,designbench2025,frontendbench2025,webbench2025,webcoderbench2026,webgenbench2025}, and interactive evaluation~\citep{artifactsbench2025,iwrbench2025,interaction2code2025}. Browser-agent environments~\citep{mind2web2023,webarena2023,weblinx2024,lu2026openclaw} add actions and temporal observations. GameDevBench~\citep{gamedevbench2026} and V-GameGym~\citep{vgamegym2025} target game development. Their targets and observation channels differ from task-defined Three.js state and physical contracts. Search-engine hot-event analyses examine Zipf-like popularity distributions~\citep{xu2018research}; such observed usage frequencies are distinct from the purpose-specific task counts in \bench.

\subsection{Agents, Programmatic Policies, and Reasoning}

\paragraph{Domain-specific agents.}
SP-Mind~\citep{yuan2026sp} coordinates tools for spatial proteomics. \emph{FinGPT-Agent}~\citep{zheng2025fingpt} targets multimodal research-report generation with task-adaptive optimization and hierarchical attention. Unlike its document outputs, \bench evaluates executable worlds and their action-conditioned behavior; neither system is an experimental baseline here.

\paragraph{Programs and environments.}
Generative code optimization represents game-playing policies as programs~\citep{kuang2025learning}; iterative optimization studies its improvement challenges~\citep{nie2026understanding}. Explicit-world-model planning~\citep{xiong2025deliberate}, diversity-driven skill evolution~\citep{xiong2026dive}, and affordance-based manipulation~\citep{liu2026palm} address agents acting in environments. Our target is generation and verification of the environment program itself. In network analysis, structural node features and clique-based search support overlapping community detection~\citep{ma2026hybrid}. Our behavioral contracts instead specify task states and action transitions rather than inferring communities.

\paragraph{Reasoning conditions.}
On-policy distillation analysis~\citep{jiang2026rocktokens}, reasoning pruning~\citep{jiang-etal-2026-drp}, attention interventions~\citep{han2026zerotuning}, and step-by-step reading~\citep{han2026read} study model improvement. Story-based memorization constraints~\citep{jiang-ferraro-2026-beyond} study evaluation conditions. These are not StateProbe components or evidence of executable-world correctness.

\subsection{Spatial and Multimodal Representations}

Point-cloud registration~\citep{liu2023regformer}, uncertainty-aware scene flow~\citep{liu2024difflow3d,liu2025difflow3d}, visual--LiDAR odometry~\citep{liu2024dvlo}, and 4D point-cloud understanding~\citep{liu2025mamba4d} estimate observed geometry and dynamics. Spatial preference optimization~\citep{shen2026fine,lu2025uni}, ELSA3D~\citep{yu2026elsa3d}, CoRe3D~\citep{yu2025core3d}, and geometrically constrained knowledge injection~\citep{wu2026injection} connect language and geometry. Medical VQA~\citep{wu2024detecting}, histopathology architecture search~\citep{su2025large}, and document annotation~\citep{han2025latex2layout} illustrate task-specific visual evidence. These capabilities complement, but do not certify, numerical or interaction-level behavior of generated code.

\subsection{Failure Transparency and Objective Trade-offs}

Failure-transparent agents~\citep{zhu2026failure} distinguish failed observations from unsupported success claims; WebShell detection~\citep{han2025can} analyzes security-sensitive behavior. Type-shift calibration~\citep{zhu2026stress} and outlier-aware adaptation~\citep{yongcanyustamp,lu2025out} address other robustness settings. Trustworthy reasoning~\citep{wang2025comprehensive}, SFT surveys~\citep{yu2026supervised}, and controlled SFT/RL comparisons~\citep{yu2026reassessing} further motivate explicit evaluation conditions.

Recommendation unlearning~\citep{li2025multi,11270220,chen2024post} and concept removal in detectors~\citep{wu2026unlearning} assess target removal separately from retained functionality. This distinction is relevant to multi-objective evaluation, but unlearning is not synonymous with preventing benchmark contamination.

\subsection{3D and Visual World Evaluation}

Visual scene benchmarks~\citep{gt23dbench2024,relscene2024,scenethesis2025,physcene2024} assess geometry and layout; physical reasoning benchmarks~\citep{intphys2018,physion2021,phyre2019} assess motion and physical prediction. WorldScore~\citep{worldscore2025} measures world controllability, quality, and dynamics, while WorldModelBench~\citep{worldmodelbench2025} evaluates video instruction following and physics adherence. CLeaR~\citep{zhou2026clear} studies reference-content leakage versus style fidelity, not hidden-test leakage. Human-perceived video fakeness~\citep{fu2025learning} provides spatially and temporally grounded appearance judgments.

StateProbe complements these output-side measures with controlled actions and runtime contracts. Mutation testing~\citep{jia2011mutation} probes checker sensitivity to documented defects; neither this calibration nor a state interface alone proves complete rendering fidelity.

\section{Dataset Details and Statistics}
\label{app:dataset_details}

\subsection{Top-Level Statistics and Domain Taxonomy}
\label{app:stats}

\begin{table}[H]
  \caption{Top-level statistics of \bench. The full benchmark contains $2{,}026$ canonical tasks; the \benchcore evaluation set used in the main results is a $205$-task hard subset selected for low reference-model $\tcov$.}
  \label{tab:stats}
  \centering\small
  \begin{tabular}{lr}
    \toprule
    Total canonical tasks            & $2{,}026$ \\
    Macro categories                 & 3 (Simulation, Rendering, Application) \\
    Fine-grained domains             & 15 \\
    Difficulty levels                & 6 (D1--D6) \\
    Asset tiers                      & 3 (Tier 0 / 1 / 2) \\
    \midrule
    \benchcore (main eval set)       & 205 tasks \\
    \quad Simulation / Rendering / Application & $63$ / $47$ / $95$ \\
    \quad Difficulty $\geq$ D5       & $164$ ($80.0\%$) \\
    \quad Tasks with \texttt{.glb} assets & $25$ ($12.2\%$) \\
    \quad Shared 3D-asset files      & $73$ \\
    \bottomrule
  \end{tabular}
\end{table}

Tables~\ref{tab:stats} and~\ref{tab:taxonomy} summarize the reported task taxonomy. Core contains $205$ difficulty-selected tasks, with $164$ at D5--D6 and $25$ requiring GLB assets. These counts do not establish a population distribution.

\begin{table}[H]
\centering
\caption{
Task taxonomy of \bench. Tasks are grouped by the primary user intent of the generated 3D world, with fine-grained domains retained for diagnostic analysis.
}
\label{tab:taxonomy}
\renewcommand{\arraystretch}{1.15}
\resizebox{\linewidth}{!}{
\begin{tabular}{l|p{3.2cm}|p{5.0cm}|p{4.2cm}|r}
\hline
\textbf{Macro Category} & \textbf{User Intent} & \textbf{Fine-Grained Domains} & \textbf{Primary Evaluation Focus} & \textbf{\#Tasks} \\
\hline
Simulation
& Observe system evolution
& \texttt{physics}, \texttt{soft\_body}, \texttt{complex\_system}, \texttt{molecular}, \texttt{chemistry}/\texttt{lab}
& Physical or chemical constraints, dynamic evolution, conservation behavior, stable state transitions
& 649 \\
\hline
Rendering
& Control visual presentation
& \texttt{shaders}, \texttt{materials}, \texttt{postprocess}, \texttt{graphics}, \texttt{creative}
& Rendering parameters, material appearance, shader behavior, post-processing effects, procedural visual states
& 620 \\
\hline
Application
& Complete an interactive goal
& \texttt{game}, \texttt{animation}, \texttt{architecture}, \texttt{product}, \texttt{visualization}
& Functional logic, interaction workflow, task completion, synchronization between user actions and world state
& 757 \\
\hline
\textbf{Total}
& --
& 15 domains
& --
& \textbf{2{,}026} \\
\hline
\end{tabular}}
\end{table}

\subsection{Splits and Difficulty}
\label{app:splits_and_difficulty}

Core, Extended, and Dev are reported canonical partitions; Robust is derived. Inputs and contracts have distinct release visibility (Table~\ref{tab:splits_appendix}). Table~\ref{tab:domain_difficulty} gives Core counts, not a general difficulty estimate.

\begin{table}[H]
\centering
\caption{Reported split accounting. Visibility refers to the inspection package; earlier public archives differ, as discussed in Appendix~\ref{app:release}.}
\label{tab:splits_appendix}
\small
\begin{tabular}{lrlll}
\toprule
Split & Size & Inputs & Contracts & Use \\
\midrule
Core & 205 & Public & Held out & Difficult-case diagnosis \\
Extended & 1,621 & Public & Held out & Broader curated comparison \\
Dev & 200 & Public & Public & Integration and debugging \\
Robust & 615 & Public & Held out & Three derived variants per Core task \\
\bottomrule
\end{tabular}
\end{table}

\begin{table}[H]
\centering
\caption{\benchcore distribution over the (domain, difficulty) grid. The right-most column is the per-domain total ($n$).}
\label{tab:domain_difficulty}
\centering\small
\begin{tabular}{ll|cccc|r}
\toprule
\textbf{Domain} & \textbf{Macro} & D3 & D4 & D5 & D6 & $n$ \\
\midrule
physics        & Sim.    & 0 & 10 & 11 & \phantom{0}5 & 26 \\
chemistry      & Sim.    & 0 & \phantom{0}2 & \phantom{0}5 & \phantom{0}4 & 11 \\
soft\_body     & Sim.    & 0 & \phantom{0}2 & \phantom{0}6 & \phantom{0}1 & \phantom{0}9 \\
molecular      & Sim.    & 0 & \phantom{0}2 & \phantom{0}5 & \phantom{0}7 & 14 \\
complex\_sys.  & Sim.    & 0 & \phantom{0}0 & \phantom{0}2 & \phantom{0}1 & \phantom{0}3 \\
\midrule
graphics       & Render. & 0 & \phantom{0}0 & \phantom{0}1 & \phantom{0}0 & \phantom{0}1 \\
materials      & Render. & 0 & \phantom{0}0 & \phantom{0}9 & \phantom{0}3 & 12 \\
creative       & Render. & 0 & \phantom{0}3 & \phantom{0}6 & \phantom{0}0 & \phantom{0}9 \\
shaders        & Render. & 0 & \phantom{0}1 & \phantom{0}5 & \phantom{0}6 & 12 \\
postprocess    & Render. & 0 & \phantom{0}0 & \phantom{0}6 & \phantom{0}7 & 13 \\
\midrule
product        & App.    & 0 & 12 & \phantom{0}5 & \phantom{0}3 & 20 \\
visualization  & App.    & 2 & \phantom{0}1 & \phantom{0}7 & \phantom{0}7 & 17 \\
game           & App.    & 0 & \phantom{0}5 & 14 & \phantom{0}4 & 23 \\
architecture   & App.    & 0 & \phantom{0}0 & \phantom{0}8 & 10 & 18 \\
animation      & App.    & 0 & \phantom{0}1 & \phantom{0}7 & \phantom{0}9 & 17 \\
\midrule
\textbf{Total} &         & \textbf{2} & \textbf{39} & \textbf{97} & \textbf{67} & \textbf{205} \\
\bottomrule
\end{tabular}
\end{table}

\subsection{Release Separation}
\label{app:antileak}

Model inputs and evaluator-side actions/checks are separate. Robust changes prompts, constants, counts, states, or assets; perturbation alone does not prove contamination resistance. Historical exposure, reconstructed payloads, and asset substitutions must be tracked by version (Appendix~\ref{app:release}).

\section{Supplementary Diagnostics}
\label{app:extended_results}

\subsection{Cost / Time Accounting and Hourly-Rate Sensitivity}
\label{app:human}

The reported RoA/TEM values use a price, latency, and human-effort model, not measured replacement labor or invoices.

\paragraph{Model cost.}
Table~\ref{tab:api_pricing} lists the rate assumptions used in the accounting. The DeepSeek-V3.2 row uses a V4 Pro price proxy, not the V3.2 billed rate. These are fixed accounting assumptions rather than current quotations or invoice-level costs, which also depend on retries, caching, reasoning-token treatment, and output selection.

\begin{table}[h]
\centering
\small
\caption{Reported input/output rate assumptions (\$/1M tokens), not current quotations. The DeepSeek-V3.2 row is a proxy.}
\label{tab:api_pricing}
\renewcommand{\arraystretch}{1.05}
\begin{tabular}{lcc}
\toprule
\textbf{Model} & \textbf{Input (\$/1M tok)} & \textbf{Output (\$/1M tok)} \\
\midrule
GPT-5.4                 & 2.50  & 15.00 \\
Gemini-3.1-Pro-Preview  & 2.00  & 12.00 \\
Claude Sonnet 4.6       & 3.00  & 15.00 \\
Claude Opus 4.6         & 5.00  & 25.00 \\
Qwen3.6-Plus            & 0.50  & \phantom{0}3.00 \\
DeepSeek-V3.2           & 0.435 & \phantom{0}0.87 \\
DeepSeek-V4-Flash       & 0.14  & \phantom{0}0.28 \\
Kimi-K2.5               & 0.60  & \phantom{0}2.00 \\
MiniMax-M2.7            & 0.30  & \phantom{0}1.20 \\
\bottomrule
\end{tabular}
\end{table}

\paragraph{Generation time.}
The accounting uses logged API durations converted to hours. Token/latency summaries in Table~\ref{tab:full_leaderboard} are reported averages, not medians or total deployment time. Evaluation and human repair overhead are not measured by TEM.

\paragraph{Human effort.}
The human-effort schedule uses seven practitioner estimates ranging from $120$ to $480$ minutes (mean $284$, median $280$). Unannotated tasks use Table~\ref{tab:human_time_fallback}. These are estimates rather than observed completion times: increasing assumed effort increases both utility proxies.

\begin{table}[h]
\centering
\small
\caption{Difficulty-conditioned fallback for $H_{\text{human}}(t)$ when a task lacks an explicit expert estimate.}
\label{tab:human_time_fallback}
\renewcommand{\arraystretch}{1.05}
\begin{tabular}{lcc}
\toprule
\textbf{Difficulty} & \textbf{Fallback (min)} & \textbf{Hours} \\
\midrule
D3 (basic interaction)               & 120 & 2 \\
D4 (multi-step interaction)          & 240 & 4 \\
D5 (physics / animation / shaders)   & 360 & 6 \\
D6 (complex system / multi-asset)    & 480 & 8 \\
\bottomrule
\end{tabular}
\end{table}

\paragraph{Wage sensitivity.}
The original reference rate is $R=\$60$/hour~\citep{wellfound_webgl_salary,talent_web_developer_salary}. A common change in $R$ rescales every RoA by $R/60$ without changing rankings; TEM is unchanged. The $25/60/100$ rates in Table~\ref{tab:roa_sensitivity} are illustrative scenarios, not verified regional or seniority categories.

\begin{table}[H]
\centering
\small
\caption{RoA (\$/\$) under three illustrative common wage assumptions. Rankings are $R$-invariant; the numerator is an estimated labor-cost proxy.}
\label{tab:roa_sensitivity}
\renewcommand{\arraystretch}{1.05}
\begin{tabular}{lrrr}
\toprule
\textbf{Model} & $R{=}\$25$ & $R{=}\$60$ \textit{(default)} & $R{=}\$100$ \\
\midrule
GPT-5.4                 & \phantom{0,0}526.9  & \phantom{0}1{,}264.6 & \phantom{0}2{,}107.7 \\
Gemini-3.1-Pro-Preview  & \phantom{0,0}646.3  & \phantom{0}1{,}551.2 & \phantom{0}2{,}585.3 \\
Claude Sonnet 4.6       & \phantom{0,0}425.3  & \phantom{0}1{,}020.7 & \phantom{0}1{,}701.2 \\
Claude Opus 4.6         & \phantom{0,0}234.3  & \phantom{0,0}562.3   & \phantom{0,0}937.2 \\
Qwen3.6-Plus            & \phantom{0}1{,}810.0 & \phantom{0}4{,}343.9 & \phantom{0}7{,}239.8 \\
DeepSeek-V3.2           & \phantom{0}6{,}058.2 & 14{,}538.8           & 24{,}231.3 \\
DeepSeek-V4-Flash       & \phantom{0}7{,}979.5 & 19{,}150.4           & 31{,}917.3 \\
Kimi-K2.5               & \phantom{0}1{,}693.0 & \phantom{0}4{,}062.7 & \phantom{0}6{,}771.2 \\
MiniMax-M2.7            & \phantom{0}5{,}134.3 & 12{,}322.2           & 20{,}537.0 \\
\bottomrule
\end{tabular}
\end{table}

\subsection{Full Model Leaderboard with System-Level Diagnostics}
\label{app:leaderboard}

Table~\ref{tab:full_leaderboard} reports Core scores, crash/probe rates over $205$ planned tasks, and average recorded generation tokens/latency. Crash rates and score rankings differ. Complete probe omission is distinct from stale or malformed state. Token counts vary by about $2.9\times$; model prices and effort assumptions also affect RoA.

\begin{table}[H]
\centering
\caption{Core results and diagnostics. Crash/probe percentages use $205$ planned tasks. Tokens and latency are recorded averages, not medians; RoA/TEM are model-based \$/\$ and hr/hr proxies.}
\label{tab:full_leaderboard}
\renewcommand{\arraystretch}{1.15}
\resizebox{\textwidth}{!}{%
\begin{tabular}{l|cccc|cc|rr|rr}
\toprule
\textbf{Model} & \textbf{V-Cov} & \textbf{A-Cov} & \textbf{S-Cov} & \textbf{T-Cov} & \textbf{Crash\%} & \textbf{Probe\%} & \textbf{\shortstack{mean\\tokens}} & \textbf{\shortstack{mean\\latency}} & \textbf{RoA} & \textbf{TEM} \\
\midrule
\multicolumn{11}{c}{\textit{Proprietary models}} \\
\midrule
GPT-5.4                & 27.8 & 66.2 & 46.1 & 26.3 & \phantom{0}3.4 & 0.5  & \phantom{0}6{,}799 & \phantom{0}88\,s & \phantom{0}1{,}264.6 & 68.87 \\
Gemini-3.1-Pro-Preview & 26.5 & 67.4 & 45.9 & 25.1 & \phantom{0}2.0 & 0.0  & \phantom{0}6{,}510 & 105\,s          & \phantom{0}1{,}551.2 & 56.81 \\
Claude Sonnet 4.6      & 18.7 & 56.5 & 33.4 & 17.0 & 13.7           & 3.4  & \phantom{0}5{,}418 & \phantom{0}79\,s & \phantom{0}1{,}020.7 & 48.34 \\
Claude Opus 4.6        & 17.5 & 58.6 & 38.3 & 15.8 & \phantom{0}7.3 & 7.3  & \phantom{0}5{,}459 & \phantom{0}81\,s & \phantom{00,}562.3 & 44.19 \\
\midrule
\multicolumn{11}{c}{\textit{Additional API-served models}} \\
\midrule
Qwen3.6-Plus           & 25.3 & 65.3 & 43.6 & 23.7 & \phantom{0}6.8 & 0.0  & \phantom{0}9{,}031 & 168\,s & \phantom{0}4{,}343.9 & 37.10 \\
DeepSeek-V3.2          & 21.9 & 60.5 & 37.4 & 20.7 & \phantom{0}9.8 & 0.0  & \phantom{0}7{,}510 & 244\,s & 14{,}538.8 & 22.49 \\
DeepSeek-V4-Flash      & 16.5 & 40.6 & 30.7 & 15.9 & \textbf{42.4}  & 0.0  & 15{,}843           & 201\,s & \textbf{19{,}150.4} & 23.23 \\
Kimi-K2.5              & 18.2 & 50.9 & 34.4 & 17.1 & 14.6           & \textbf{10.2} & \phantom{0}6{,}488 & 153\,s & 4,062.7     & 38.26   \\
MiniMax-M2.7           & 23.0 & 57.6 & 39.9 & 21.7 & 14.6           & 0.0  & 10{,}244           & 191\,s & 12,322.2     & 48.36   \\
\bottomrule
\end{tabular}}
\end{table}

Table~\ref{tab:leaderboard_domains} gives domain scores. Small or uneven domain samples limit interpretation; the rows are descriptive diagnostics, not population-wide estimates of task difficulty.

\begin{table}[H]
\centering
\caption{Per-domain V-Cov ($\%$) on \benchcore for all nine evaluated models. The number of \benchcore tasks in each domain ($n$) is given in the rightmost column. Every model collapses on \texttt{creative}, and \texttt{graphics} carries only one task and is reported for completeness.}
\label{tab:leaderboard_domains}
\renewcommand{\arraystretch}{1.15}
\resizebox{\textwidth}{!}{%
\begin{tabular}{ll|ccccccccc|r}
\toprule
\textbf{Domain} & \textbf{Macro} & \textbf{GPT-5.4} & \textbf{Gem.\,3.1} & \textbf{Sonnet} & \textbf{Opus} & \textbf{Qwen} & \textbf{DS-V3.2} & \textbf{DS-V4} & \textbf{Kimi} & \textbf{MiniMax} & $n$ \\
\midrule
physics        & Sim.    & 21.7 & 28.0 & 18.2 & 15.9 & 21.7 & 21.4 & 16.6 & 11.9 & 23.0 & 26 \\
chemistry      & Sim.    & 36.1 & 30.9 & 31.0 & 23.4 & 23.9 & 29.3 & \phantom{0}7.8 & 19.6 & 17.0 & 11 \\
soft\_body     & Sim.    & 32.3 & 27.7 & 24.3 & 21.1 & 26.1 & 32.3 & \phantom{0}5.0 & 25.7 & 21.8 & \phantom{0}9 \\
molecular      & Sim.    & 28.0 & 24.0 & 20.8 & 17.7 & 31.9 & 11.3 & 13.2 & \phantom{0}6.0 & 29.3 & 14 \\
complex\_sys.  & Sim.    & 15.7 & 15.7 & \phantom{0}9.3 & 13.0 & 13.0 & \phantom{0}3.7 & \phantom{0}0.0 & \phantom{0}5.6 & 10.2 & \phantom{0}3 \\
\midrule
graphics       & Render. & \phantom{0}0.0 & \phantom{0}9.1 & \phantom{0}0.0 & \phantom{0}9.1 & \phantom{0}0.0 & \phantom{0}0.0 & \phantom{0}0.0 & \phantom{0}9.1 & \phantom{0}0.0 & \phantom{0}1 \\
materials      & Render. & 54.8 & 35.5 & 30.6 & 28.3 & 41.6 & 32.6 & 37.8 & 30.3 & 45.8 & 12 \\
creative       & Render. & \phantom{0}0.0 & \phantom{0}0.0 & \phantom{0}0.0 & \phantom{0}0.0 & \phantom{0}0.0 & \phantom{0}0.0 & \phantom{0}0.0 & \phantom{0}0.0 & \phantom{0}0.0 & \phantom{0}9 \\
shaders        & Render. & 49.2 & 60.5 & 25.8 & 28.2 & 42.3 & 38.4 & 25.3 & 42.8 & 41.1 & 12 \\
postprocess    & Render. & 40.8 & 40.7 & 38.6 & 27.3 & 32.4 & 38.6 & 29.1 & 37.6 & 28.8 & 13 \\
\midrule
product        & App.    & 14.9 & 14.8 & \phantom{0}7.6 & \phantom{0}5.4 & 11.9 & 10.0 & \phantom{0}4.9 & \phantom{0}6.5 & 10.3 & 20 \\
visualization  & App.    & 37.7 & 30.4 & 21.9 & 27.1 & 27.0 & 31.7 & 27.3 & 27.0 & 32.2 & 17 \\
game           & App.    & \phantom{0}9.7 & \phantom{0}7.0 & \phantom{0}6.1 & \phantom{0}8.0 & 12.8 & \phantom{0}7.5 & \phantom{0}8.2 & \phantom{0}5.0 & \phantom{0}8.7 & 23 \\
architecture   & App.    & 29.1 & 37.3 & 15.8 & 19.7 & 36.3 & 23.2 & 28.7 & 28.1 & 31.0 & 18 \\
animation      & App.    & 32.4 & 22.9 & 20.5 & 16.4 & 33.1 & 26.9 & 13.9 & 17.1 & 21.2 & 17 \\
\midrule
\textbf{Overall} & --    & \textbf{27.8} & \textbf{26.5} & \textbf{18.7} & \textbf{17.5} & \textbf{25.3} & \textbf{21.9} & \textbf{16.5} & \textbf{18.2} & \textbf{23.0} & \textbf{205} \\
\bottomrule
\end{tabular}}
\end{table}

\subsection{Core--Extended Utility Comparison}
\label{app:extended_utility}

\begin{table}[!htbp]
\centering
\small
\caption{Core and Extended utility proxies. RoA is in \$/\$ and TEM in hr/hr; these are accounting estimates, not measured savings.}
\label{tab:extended_utility}
\begin{tabular}{lrrrr}
\toprule
\multirow{2}{*}{\textbf{Model}} & \multicolumn{2}{c}{\textbf{RoA}} & \multicolumn{2}{c}{\textbf{TEM}} \\
\cmidrule(lr){2-3}\cmidrule(lr){4-5}
 & \textbf{Core} & \textbf{Extended} & \textbf{Core} & \textbf{Extended} \\
\midrule
GPT-5.4 & 1,264.6 & 1,687.4 & 68.87 & 61.92 \\
Gemini 3.1 Pro Preview & 1,551.2 & 1,438.6 & 56.81 & 82.37 \\
Claude Sonnet 4.6 & 1,020.7 & 1,596.3 & 48.34 & 54.86 \\
Claude Opus 4.6 & 562.3 & 746.8 & 44.19 & 70.41 \\
Qwen3.6-Plus & 4,343.9 & 6,183.7 & 37.10 & 43.28 \\
DeepSeek-V3.2 & 14,538.8 & 12,764.9 & 22.49 & 28.73 \\
DeepSeek-V4-Flash & 19,150.4 & 28,615.3 & 23.23 & 21.84 \\
Kimi-K2.5 & 4,062.7 & 4,681.5 & 38.26 & 56.07 \\
MiniMax-M2.7 & 12,322.2 & 17,426.8 & 48.36 & 66.35 \\
\midrule
Nine-model mean & 6,535.2 & 8,349.0 & 43.07 & 53.98 \\
\bottomrule
\end{tabular}
\end{table}

The nine-model averages rise from $6{,}535.2$ to $8{,}349.0$ for RoA and from $43.07$ to $53.98$ for TEM. Cross-split Spearman rank correlations are $0.93$ and $0.78$, respectively. DeepSeek-V4-Flash retains the highest RoA, while Gemini leads Extended TEM, illustrating distinct cost and speed orderings. These comparisons use the accounting assumptions in Appendix~\ref{app:human}, rather than measured developer savings.

\subsection{Per-Domain Economic Profile (All Seven Models)}
\label{app:roa_tem_full}

\begin{figure}[H]
  \centering
  \includegraphics[width=\linewidth]{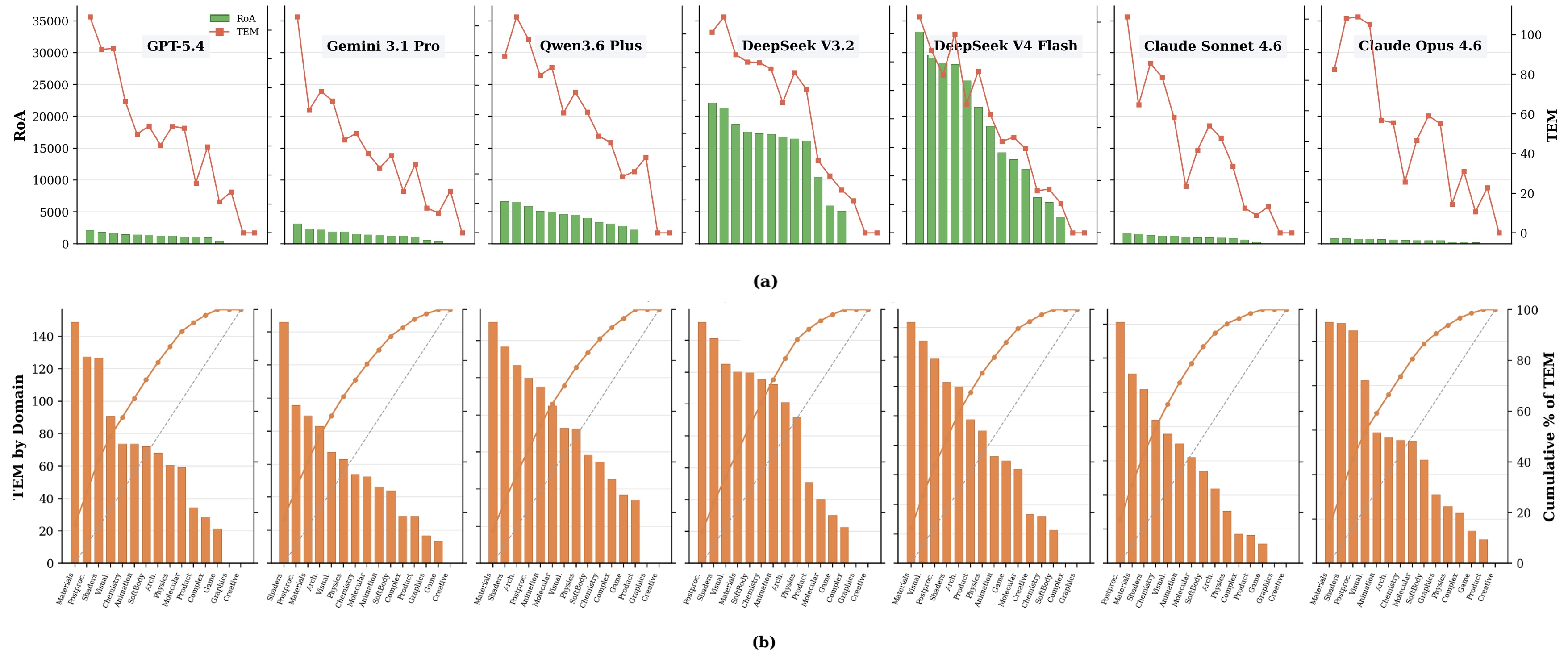}
  \caption{
    Per-domain economic profiles for a seven-model Core subset. Layout matches Figure~\ref{fig:roa_tem_pareto}: row (a) plots per-domain RoA (green bars) with per-domain TEM (red line); row (b) plots per-domain TEM (orange bars) with cumulative TEM share (line, right axis). The larger shared RoA axis accommodates the DeepSeek panels.
  }
  \label{fig:roa_tem_pareto_full}
\end{figure}

Figure~\ref{fig:roa_tem_pareto_full} shows domain-level cost/time proxies for the seven-model subset with economic-profile plots, under the accounting assumptions in Appendix~\ref{app:human}.

\subsection{Per-Model Paradigm Comparison}
\label{app:paradigm_per_model}

Table~\ref{tab:paradigm_per_model} comes from a separate seven-model comparison archive. Its reference column differs from the primary Core table, including the DeepSeek-V4-Flash entry, and its VLM averages are source-judge scores. We report each archive separately rather than pooling their reference scores. Per-pair and per-model correlations also use different aggregation units.

\begin{table}[H]
\centering
\caption{A separate seven-model comparison archive. The archived reference score is not interchangeable with the primary Core score; VLM averages here are source-judge scores, not screenshot pass rates. $n$ is the per-model sample size; $\tau_b$ uses the displayed averages.}
\label{tab:paradigm_per_model}
\small
\renewcommand{\arraystretch}{1.15}
\begin{tabular}{lrrrrr}
\toprule
\textbf{Model} & \textbf{Archive ref.} & \textbf{DOM avg.} & \textbf{VLM avg.} & \textbf{Agent avg.} & \textbf{n} \\
\midrule
GPT-5.4                       &  27.8 &  50.0 &  46.7 &  49.6 & 205 \\
Gemini-3.1-Pro                &  26.5 &  47.6 &  40.2 &  49.2 & 205 \\
Qwen3.6-Plus                  &  25.4 &  52.0 &  34.2 &  47.5 & 204 \\
DeepSeek-V3.2                 &  21.9 &  54.7 &  33.2 &  46.2 & 205 \\
DeepSeek-V4-Flash             &  19.9 &  51.1 &  35.1 &  51.8 & 205 \\
Claude-Sonnet-4.6             &  18.7 &  51.1 &  36.5 &  50.8 & 205 \\
Claude-Opus-4.6               &  17.5 &  51.8 &  36.8 &  48.1 & 205 \\
\midrule
\textit{Per-model Kendall $\tau_b$} & --- & $-0.293$ & $+0.143$ & $+0.048$ & --- \\
\bottomrule
\end{tabular}
\end{table}

\section{Detailed Error Analysis and Case Studies}
\label{app:detailed_errors}

\subsection{Diagnostic Denominators}
\label{app:errors_full}

Tables~\ref{tab:failures_per_model} and~\ref{tab:failures_per_domain} summarize diagnostic labels. Behavioral labels can co-occur, so row sums can exceed $100\%$; crash/probe percentages use all planned tasks. The domain table lists a subset of the taxonomy with its own reported $n$. Primary-summary shares, multi-label prevalence, and global failure counts have different denominators and should be interpreted separately.

\begin{table}[H]
\centering
\caption{Archived per-model diagnostics. Behavioral percentages use zero-T-Cov records ($N$) and are non-exclusive; crash/probe rates use all $205$ planned tasks. Behavioral columns need not sum to $100\%$.}
\label{tab:failures_per_model}
\renewcommand{\arraystretch}{1.15}
\resizebox{\textwidth}{!}{%
\begin{tabular}{l|ccccc|rr|r}
\toprule
\textbf{Model} & \textbf{Drift} & \textbf{Phys.} & \textbf{Chain} & \textbf{UI} & \textbf{Sem.} & \textbf{Crash\%} & \textbf{Probe\%} & \textbf{$N$} \\
\midrule
GPT-5.4                  & 97 & 0 & 92 & 20 & 0 & \phantom{0}3.4           & \phantom{0}0.5            & 71 \\
Gemini-3.1-Pro-Preview   & 93 & 0 & 92 & 27 & 0 & \textbf{\phantom{0}2.0}  & \phantom{0}0.0            & 71 \\
Claude Sonnet 4.6        & 97 & 1 & 88 & 18 & 0 & 13.7                     & \phantom{0}3.4            & 76 \\
Claude Opus 4.6          & 93 & 1 & 92 & 24 & 0 & \phantom{0}7.3           & \phantom{0}7.3            & 75 \\
Qwen3.6-Plus             & 98 & 0 & 94 & 19 & 2 & \phantom{0}6.8           & \phantom{0}0.0            & 62 \\
DeepSeek-V3.2            & 96 & 0 & 88 & 28 & 1 & \phantom{0}9.8           & \phantom{0}0.0            & 78 \\
DeepSeek-V4-Flash        & 95 & 5 & 91 & 23 & 0 & \textbf{42.4}            & \phantom{0}0.0            & 44 \\
Kimi-K2.5                & 95 & 2 & 92 & 23 & 0 & 14.6                     & \textbf{10.2}             & 60 \\
MiniMax-M2.7             & 97 & 0 & 98 & 22 & 0 & 14.6                     & \phantom{0}0.0            & 59 \\
\midrule
\textbf{Overall}         & \textbf{96} & \textbf{1} & \textbf{92} & \textbf{23} & \textbf{0} & \textbf{12.8} & \textbf{2.3} & \textbf{596} \\
\bottomrule
\end{tabular}}
\end{table}

\begin{table}[H]
\centering
\caption{Archived domain diagnostics with reported per-domain $n$. Several rows exceed $100\%$; columns are not a mutually exclusive primary-cause distribution. Omitted domains and different label scopes preclude pooling with the global summaries.}
\label{tab:failures_per_domain}
\renewcommand{\arraystretch}{1.15}
\centering\small
\begin{tabular}{ll|ccccc|r}
\toprule
\textbf{Domain} & \textbf{Macro} & \textbf{Drift} & \textbf{Phys.} & \textbf{Chain} & \textbf{UI} & \textbf{Sem.} & $n$ \\
\midrule
physics        & Sim.    & 90 & 0 & \phantom{0}9 & \phantom{0}3 & 0 & \phantom{0}99 \\
chemistry      & Sim.    & 70 & 0 & 39 & 35 & 0 & \phantom{0}23 \\
soft\_body     & Sim.    & 57 & 0 & 29 & 36 & 0 & \phantom{0}14 \\
molecular      & Sim.    & 69 & 0 & 10 & 41 & 0 & \phantom{0}29 \\
complex\_sys.  & Sim.    & 100 & 0 & \phantom{0}0 & \phantom{0}0 & 0 & \phantom{00}7 \\
\midrule
creative       & Render. & 89 & 0 & 12 & \phantom{0}1 & 0 & \phantom{0}75 \\
\midrule
product        & App.    & 77 & 0 & 25 & 10 & 0 & \phantom{0}96 \\
visualization  & App.    & 96 & 0 & \phantom{0}4 & 22 & 0 & \phantom{0}45 \\
game           & App.    & 66 & 0 & 25 & 14 & 0 & 118 \\
architecture   & App.    & 85 & 0 & \phantom{0}9 & 18 & 0 & \phantom{0}34 \\
animation      & App.    & 64 & 0 & 26 & 21 & 0 & \phantom{0}42 \\
\bottomrule
\end{tabular}
\end{table}

\subsection{Representative Case Walk-through}
\label{app:case_study}

\begin{figure}[H]
  \centering
  \includegraphics[width=\linewidth]{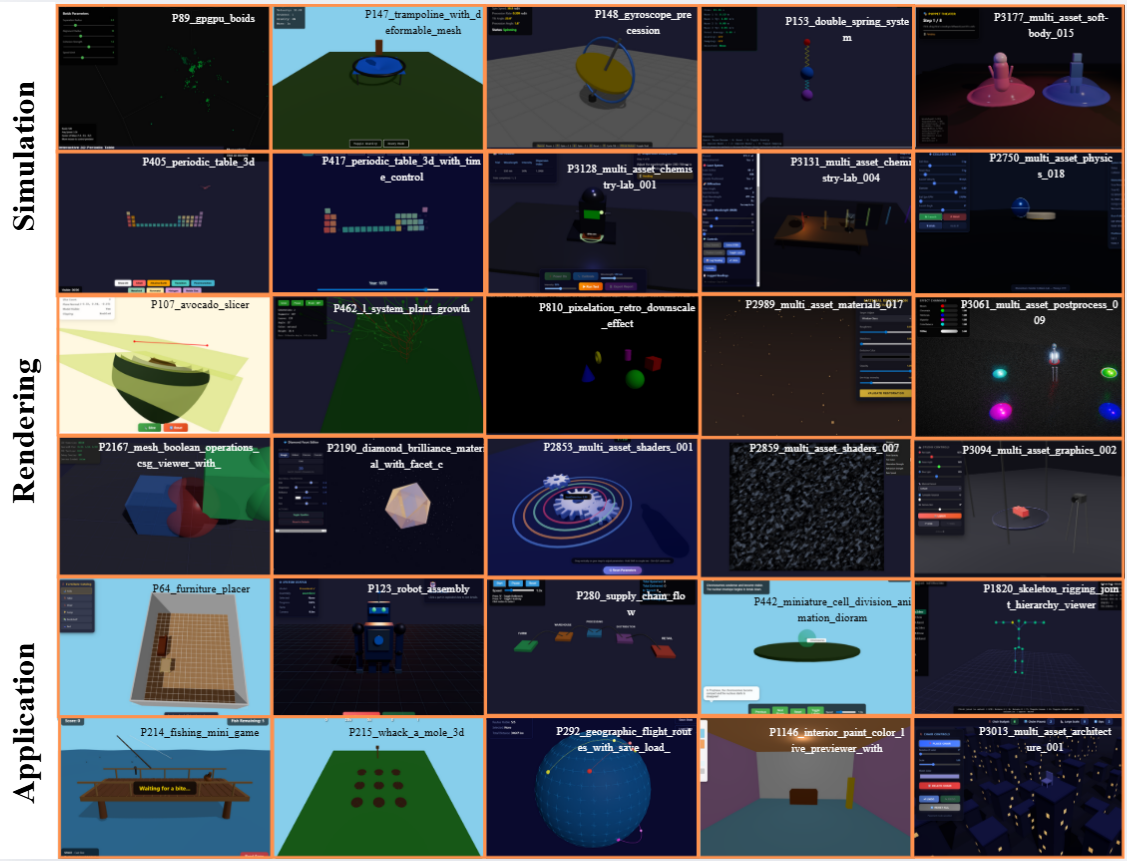}
  \caption{Six representative \benchcore tasks (two per macro-category).}
  \label{fig:cases}
\end{figure}

Figure~\ref{fig:cases} illustrates six archived Core examples. Listing~\ref{lst:example_case} preserves an abridged basketball prompt and its visible state schema. Archive IDs are not assumed to match current release IDs; the prompt is not the hidden executable rubric.

\begin{lstlisting}[language={},caption={Abridged \texttt{task.json} for P253: Basketball Free Throw with Particle Effects.},label={lst:example_case},aboveskip=5pt]
"id": "P253_basketball_free_throw_with_particle_effe",
"title": "Basketball Free Throw with Particle Effects",
"domain": "game",
"difficulty": "L5",
"framework": "three.js",
"prompt":"Create a single HTML file with an interactive 3D basketball free throw game using Three.js. Load the 'BarramundiFish.glb' model using GLTFLoader as a trophy/mascot that sits on the scoreboard and animates when the player scores.

SCENE SETUP:
- Use a PerspectiveCamera at position (0, 5, 15) looking toward the hoop.
- Add ambient light (intensity 0.5) and a directional light at (10, 20, 10) with intensity 1.0 and castShadow enabled.
- Create a basketball court floor: a large green/brown plane at y=0 with receiveShadow.
- Create a basketball hoop: an orange torus (ring) at position (0, 8, -8) with inner radius 0.6, outer radius 0.08, oriented horizontally (rotated PI/2 on X). Add a backboard behind it as a white box (2.5 x 2 x 0.1) at (0, 9, -8.5). Add a pole as a thin grey cylinder from the ground to the backboard.
- Create a basketball: an orange sphere with radius 0.35 at starting position (0, 3, 10). Add dark curved lines texture or pattern if possible.
- Load 'BarramundiFish.glb' using GLTFLoader. Scale it to (2, 2, 2) and place it at (4, 10.5, -8.5) on top of the backboard as a quirky mascot trophy. Store reference as window.__fishModel.

PHYSICS:
- Implement simple projectile physics with gravity = $-9.81 \mathrm{m/s^2}$. Use a fixed timestep of 1/60.
- When the ball is launched, apply initial velocity based on power and angle.
- Ball velocity: vx based on aimX, vy = power * sin(angle), vz = -power * cos(angle).
- Detect collision with the hoop ring: if ball passes through the torus center (within 0.6 radius horizontally and within 0.3 vertically of the hoop center y=8, z=-8), count as a SCORE.
- Detect collision with backboard: if ball hits the backboard box, reflect the z-velocity (multiply by -0.5 for energy loss).
- Detect floor collision: if ball.y <= 0.35, bounce with vy *= -0.4, reduce horizontal velocities by 0.8.
- Ball stops when speed < 0.1 after bouncing.

PARTICLE EFFECTS:
- On SCORE: emit 100 gold/yellow particles from the hoop position. Particles spread outward with random velocities, fade out over 2 seconds, and have gravity. Use THREE.Points with BufferGeometry.
- On ball launch: emit 20 white smoke particles from the ball's starting position that fade over 0.5 seconds.
- Particles should have size attenuation and transparency.

USER INTERACTIONS:
- POWER: Press and hold SPACEBAR to charge power (0 to 20 over 2 seconds). A power bar HUD element shows the current charge. Release to lock power.
- ANGLE: After power is set, use UP/DOWN arrow keys to adjust launch angle between 30 and 75 degrees. Display angle in HUD.
- AIM: Use LEFT/RIGHT arrow keys to adjust horizontal aim (aimX) between -3 and 3. Display in HUD.
- LAUNCH: Press ENTER to launch the ball with current power, angle, and aim.
- RESET: Press 'R' to reset the ball to starting position for another throw.
- The fish trophy should rotate (spin on Y axis) and bounce up/down for 2 seconds when a score happens.

HUD/UI:
- Top-left: Score counter (div id='scoreDisplay') showing 'Score: X'
- Top-right: Shots taken counter (div id='shotsDisplay') showing 'Shots: X'
- Bottom-center: Power bar (div id='powerBar') with inner fill div (id='powerFill'), width proportional to power.
- Below power bar: Angle display (div id='angleDisplay') showing 'Angle: XX$^\circ$'
- Below angle: Aim display (div id='aimDisplay') showing 'Aim: X.X'
- Center: Status text (div id='statusText') showing current phase: 'Hold SPACE to charge', 'Use UP/DOWN for angle', 'Press ENTER to launch', 'SCORE!', or 'Miss! Press R to reset'.

STATE MANAGEMENT - Expose window.__3D_STATE__ with:
- phase: 'charging' | 'aiming' | 'ready' | 'flying' | 'scored' | 'missed' | 'idle'
- power: number (0-20)
- angle: number (30-75)
- aimX: number (-3 to 3)
- score: number (total baskets made)
- shots: number (total shots taken)
- ballPosition: {x, y, z}
- ballVelocity: {x, y, z}
- isCharging: boolean
- particleCount: number (active particles)
- fishLoaded: boolean
- fishCelebrating: boolean
- ballSpeed: number (magnitude of velocity)
- lastShotResult: 'none' | 'score' | 'miss'

Initialize phase as 'idle', power 0, angle 45, aimX 0, score 0, shots 0.",
"assets": [
    "BarramundiFish.glb"
  ],
"physics_constraints": "Projectile motion with gravity=-9.81. Floor bounce at y=0.35 with restitution 0.4. Backboard reflection with 0.5 energy loss. Ball stops when speed < 0.1. Hoop scoring detection within 0.6 radius of hoop center."
\end{lstlisting}

The prompt specifies shot phases, projectile variables, and synchronized score fields. Testing these requirements needs actions and state checks; a visible schema alone does not establish their correctness.

\begin{figure}[H]
  \centering
  \includegraphics[width=\linewidth]{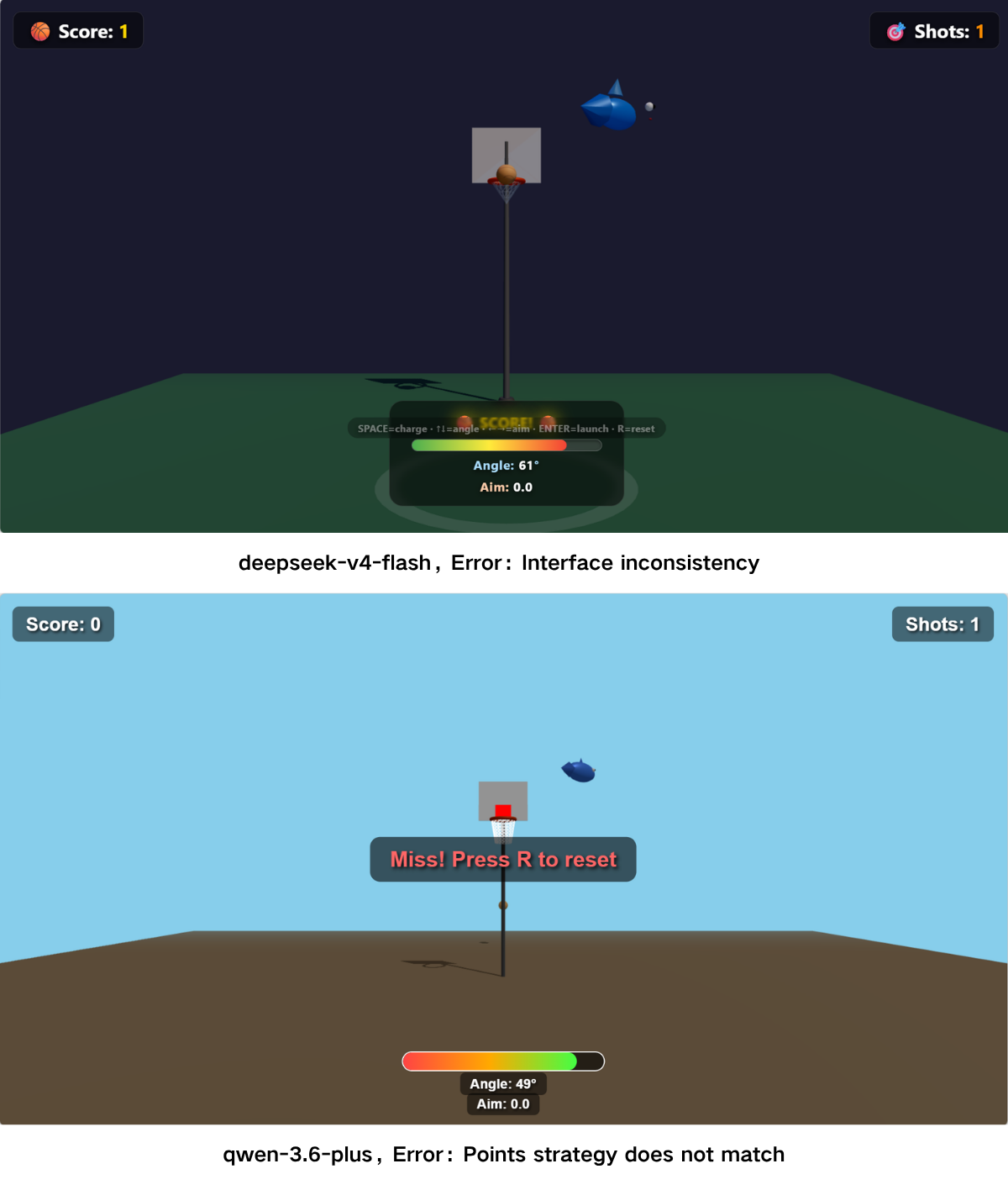}
  \caption{Basketball-generation examples with DeepSeek interface and Qwen scoring annotations.}
  \label{fig:in_case}
\end{figure}

\begin{figure}[H]
  \centering
  \includegraphics[width=0.9\linewidth]{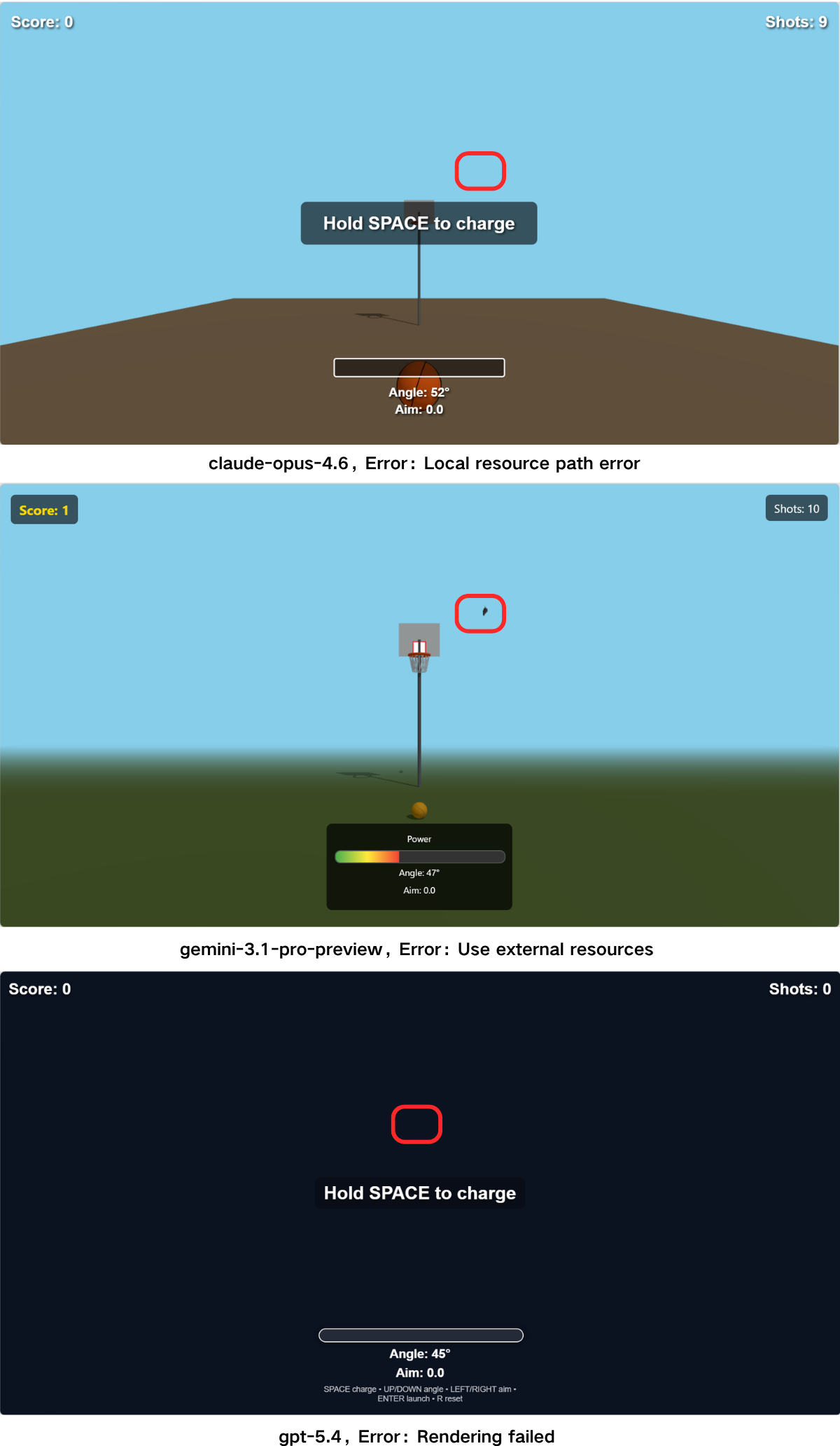}
  \caption{Basketball-generation examples with resource and rendering annotations for Opus, Gemini, and GPT.}
  \label{fig:out_case}
\end{figure}

Figures~\ref{fig:out_case} and~\ref{fig:in_case} illustrate basketball-generation outputs. Their annotations identify resource, rendering, interface, and scoring issues:
\begin{itemize}[nosep]
  \item \textbf{Resource access:} Opus uses a mismatched local path; Gemini requests an external asset. An external request can conflict with offline policy, but does not prove a sandbox escape.
  \item \textbf{Rendering and observability:} the GPT capture lacks the expected scene; the DeepSeek example is labelled as an interface inconsistency. The basketball prompt explicitly requires the fish mascot and its \texttt{window.\_\_fishModel} reference.
  \item \textbf{Scoring:} the Qwen annotation identifies a mismatch in scoring behavior. A screenshot illustrates the output, while numerical or temporal violations require action traces and contract checks.
\end{itemize}
These examples illustrate individual diagnostics rather than population failure rates.

\section{Broader Impact}
\label{app:impact}

The benchmark can help diagnose physics, interaction, and synchronization failures before generated 3D programs are deployed. Its LLM-assisted task construction and expert curation support systematic evaluation, while its effort estimates offer a transparent basis for comparing cost and time assumptions.

Plausible simulations can mislead users, and untrusted generated code can create security risks. Offline request filtering is not a guarantee of host isolation or prevention of exfiltration. Runs should exclude credentials and sensitive mounts and use explicit network/resource controls. Third-party assets retain source-specific terms; the benchmark license does not relicense them. We do not certify visual realism, safety, or production readiness from contract coverage alone.

\section{Protocol, Release, and Reproducibility Notes}
\label{app:protocol_notes}

\subsection{Score Version and Coverage Boundaries}
\label{app:metric_policy}

The Core runner computes each task's V-Cov from full-transition pass rates within highest-check tiers, weighted $0.2/0.3/0.5$ and normalized over nonempty tiers; the leaderboard averages over tasks. The $205$-task, nine-model report archive matches the rounded Core scores. The public Dev evaluator uses an unweighted individual-assertion ratio instead. These score versions are distinct and should not be compared as identical metrics.

T-Cov divides passed transitions by the contract's transition count. The Core V-Cov tier counts can omit a transition that raises an execution error before its checks are recorded; thus its denominator need not equal all scheduled transitions. Empty tiers are omitted; a task with no contributing tier receives zero. Affordance compatibility defaults can credit a missing check, and empty affordance/state sets have legacy defaults. Consequently, A-Cov is a contract diagnostic, not independent confirmation of visible object existence. Comparisons across score policies require version-matched rescoring.

Public Dev design notes also identify unconditional fail-closed affordance checks. Such a check reports unavailable verification, not physical absence. Neither a self-reported existence flag nor a particle-emission counter independently verifies rendering. Schema additions, eligibility changes, or repaired checks change the task/metric version and require paired reassessment.

\subsection{Extended and Interface-Completion Results}
\label{app:review_results}

Tables~\ref{tab:extended_results}, \ref{tab:interface_completion}, and~\ref{tab:extended_utility} report the additional evaluations conducted during review. Extended used the same evaluation settings; the interface ablation completed omitted runtime interfaces before re-evaluation. The public package does not include the complete Extended run ledger or paired interface patches. The ablation concerns interface completion, not arbitrary behavioral repair. Core and Extended are curated splits, so their rank agreement characterizes these splits rather than a population-wide distribution.

\subsection{Runtime Interface and Rendering Audits}
\label{app:interface_limits}

Core reports record complete probe absence for GPT-5.4 ($1/205$), Claude Opus 4.6 ($15/205$, $7.3\%$), and Kimi-K2.5 ($21/205$, $10.2\%$). This is distinct from malformed fields or stale updates. Assigning full task credit to GPT's sole omitted probe could raise mean V-Cov by at most $0.5$ pp; this is an upper bound, not a measured repair gain. Table~\ref{tab:interface_completion} reports the completion ablation. Observability-only repair means exposing existing state without adding task logic or private expected values.

Among the $410$ GPT/Gemini Core records, the strict full-pass rule selects nine GPT and eight Gemini outputs covering twelve tasks. Passing the scheduled state checks does not alone establish scene/pixel fidelity: rendering verification additionally requires synchronized visual observations. State-transition coverage and visual fidelity are complementary measures.

\subsection{Public Artifacts and Provenance}
\label{app:release}

The Core prompt/asset snapshot is hosted at \url{https://huggingface.co/datasets/shuolucs/WorldCoder-Bench}, with evaluator code at \url{https://github.com/shuolucs/WorldCoder-Bench}. Release snapshots differ: the earlier repository documents a $2{,}004$-task source snapshot and five Dev rubric examples; the inspection package at \url{https://anonymous.4open.science/r/WorldCoder-Bench/} documents $2{,}641$ task specifications including Robust variants, plus public Dev rubrics. Its membership notes identify reconstructed records and twelve replacements by controlled augmentations. These release payloads should be distinguished from the historical evaluation payloads.

Evaluation and release IDs, task/asset/contract hashes, substitutions, and reconstructions are version-specific. Holding contracts out of generation requests establishes inference-time separation, not a guarantee against prior public exposure of historical material.

The HF Core snapshot and inspection package declare different benchmark licenses; users should consult the terms of the exact snapshot. Third-party assets retain their own source-specific terms and are not relicensed by the benchmark. Release membership and license metadata must therefore be checked at the version level.

\subsection{Reproducibility and Safety Scope}
\label{app:reproducibility}

The release provides a Dev CLI, including \texttt{code/run\_eval.sh}, for user-supplied HTML and public task/rubric pairs. The generation program sets \texttt{temperature=0.2} and \texttt{max\_tokens=32000}, without a fixed seed; other sampling settings follow the respective API defaults. Exact replay additionally depends on model/API versions, retry and output-selection policies, task/contract hashes, evaluator and browser revisions, and cost accounting. The public Dev package supports local evaluation but does not include every frozen request, hidden contract, or run-level cost ledger for the reported experiments.

The documented offline mode serves a local runtime and blocks unmatched HTTP(S). It reduces external requests but does not guarantee operating-system isolation, especially when Chromium sandbox flags are disabled. Untrusted programs should run on isolated hosts or in containers without secrets or sensitive mounts, with explicit networking policy and resource limits.

\clearpage
\section*{NeurIPS Paper Checklist}

\begin{enumerate}

\item {\bf Claims}
    \item[] Question: Do the main claims made in the abstract and introduction accurately reflect the paper's contributions and scope?
    \item[] Answer: \answerYes{}
    \item[] Justification: Claims are restricted to curated splits, recorded score policies, and tested behavior (Sections~\ref{sec:intro} and~\ref{sec:experiments}).
    \item[] Guidelines:
    \begin{itemize}
        \item The answer \answerNA{} means that the abstract and introduction do not include the claims made in the paper.
        \item The abstract and/or introduction should clearly state the claims made, including the contributions made in the paper and important assumptions and limitations. A \answerNo{} or \answerNA{} answer to this question will not be perceived well by the reviewers. 
        \item The claims made should match theoretical and experimental results, and reflect how much the results can be expected to generalize to other settings. 
        \item It is fine to include aspirational goals as motivation as long as it is clear that these goals are not attained by the paper. 
    \end{itemize}

\item {\bf Limitations}
    \item[] Question: Does the paper discuss the limitations of the work performed by the authors?
    \item[] Answer: \answerYes{}
    \item[] Justification: Appendices~\ref{app:protocol_notes} and~\ref{app:human} describe selection, observability, scoring, release, and utility limitations.
    \item[] Guidelines:
    \begin{itemize}
        \item The answer \answerNA{} means that the paper has no limitation while the answer \answerNo{} means that the paper has limitations, but those are not discussed in the paper. 
        \item The authors are encouraged to create a separate ``Limitations'' section in their paper.
        \item The paper should point out any strong assumptions and how robust the results are to violations of these assumptions (e.g., independence assumptions, noiseless settings, model well-specification, asymptotic approximations only holding locally). The authors should reflect on how these assumptions might be violated in practice and what the implications would be.
        \item The authors should reflect on the scope of the claims made, e.g., if the approach was only tested on a few datasets or with a few runs. In general, empirical results often depend on implicit assumptions, which should be articulated.
        \item The authors should reflect on the factors that influence the performance of the approach. For example, a facial recognition algorithm may perform poorly when image resolution is low or images are taken in low lighting. Or a speech-to-text system might not be used reliably to provide closed captions for online lectures because it fails to handle technical jargon.
        \item The authors should discuss the computational efficiency of the proposed algorithms and how they scale with dataset size.
        \item If applicable, the authors should discuss possible limitations of their approach to address problems of privacy and fairness.
        \item While the authors might fear that complete honesty about limitations might be used by reviewers as grounds for rejection, a worse outcome might be that reviewers discover limitations that aren't acknowledged in the paper. The authors should use their best judgment and recognize that individual actions in favor of transparency play an important role in developing norms that preserve the integrity of the community. Reviewers will be specifically instructed to not penalize honesty concerning limitations.
    \end{itemize}

\item {\bf Theory assumptions and proofs}
    \item[] Question: For each theoretical result, does the paper provide the full set of assumptions and a complete (and correct) proof?
    \item[] Answer: \answerNA{}
    \item[] Justification: No theoretical result is claimed; the equations define scores and accounting assumptions.
    \item[] Guidelines:
    \begin{itemize}
        \item The answer \answerNA{} means that the paper does not include theoretical results. 
        \item All the theorems, formulas, and proofs in the paper should be numbered and cross-referenced.
        \item All assumptions should be clearly stated or referenced in the statement of any theorems.
        \item The proofs can either appear in the main paper or the supplemental material, but if they appear in the supplemental material, the authors are encouraged to provide a short proof sketch to provide intuition. 
        \item Inversely, any informal proof provided in the core of the paper should be complemented by formal proofs provided in appendix or supplemental material.
        \item Theorems and Lemmas that the proof relies upon should be properly referenced. 
    \end{itemize}

    \item {\bf Experimental result reproducibility}
    \item[] Question: Does the paper fully disclose all the information needed to reproduce the main experimental results of the paper to the extent that it affects the main claims and/or conclusions of the paper (regardless of whether the code and data are provided or not)?
    \item[] Answer: \answerNo{}
    \item[] Justification: The paper and public package do not include every frozen request, contract/evaluator version, or run-level cost ledger (Appendix~\ref{app:reproducibility}).
    \item[] Guidelines:
    \begin{itemize}
        \item The answer \answerNA{} means that the paper does not include experiments.
        \item If the paper includes experiments, a \answerNo{} answer to this question will not be perceived well by the reviewers: Making the paper reproducible is important, regardless of whether the code and data are provided or not.
        \item If the contribution is a dataset and\slash or model, the authors should describe the steps taken to make their results reproducible or verifiable. 
        \item Depending on the contribution, reproducibility can be accomplished in various ways. For example, if the contribution is a novel architecture, describing the architecture fully might suffice, or if the contribution is a specific model and empirical evaluation, it may be necessary to either make it possible for others to replicate the model with the same dataset, or provide access to the model. In general. releasing code and data is often one good way to accomplish this, but reproducibility can also be provided via detailed instructions for how to replicate the results, access to a hosted model (e.g., in the case of a large language model), releasing of a model checkpoint, or other means that are appropriate to the research performed.
        \item While NeurIPS does not require releasing code, the conference does require all submissions to provide some reasonable avenue for reproducibility, which may depend on the nature of the contribution. For example
        \begin{enumerate}
            \item If the contribution is primarily a new algorithm, the paper should make it clear how to reproduce that algorithm.
            \item If the contribution is primarily a new model architecture, the paper should describe the architecture clearly and fully.
            \item If the contribution is a new model (e.g., a large language model), then there should either be a way to access this model for reproducing the results or a way to reproduce the model (e.g., with an open-source dataset or instructions for how to construct the dataset).
            \item We recognize that reproducibility may be tricky in some cases, in which case authors are welcome to describe the particular way they provide for reproducibility. In the case of closed-source models, it may be that access to the model is limited in some way (e.g., to registered users), but it should be possible for other researchers to have some path to reproducing or verifying the results.
        \end{enumerate}
    \end{itemize}

\item {\bf Open access to data and code}
    \item[] Question: Does the paper provide open access to the data and code, with sufficient instructions to faithfully reproduce the main experimental results, as described in supplemental material?
    \item[] Answer: \answerNo{}
    \item[] Justification: Public inputs and evaluator code are linked, but the complete frozen material needed for the main results is not released (Appendix~\ref{app:release}).
    \item[] Guidelines:
    \begin{itemize}
        \item The answer \answerNA{} means that paper does not include experiments requiring code.
        \item Please see the NeurIPS code and data submission guidelines (\url{https://neurips.cc/public/guides/CodeSubmissionPolicy}) for more details.
        \item While we encourage the release of code and data, we understand that this might not be possible, so \answerNo{} is an acceptable answer. Papers cannot be rejected simply for not including code, unless this is central to the contribution (e.g., for a new open-source benchmark).
        \item The instructions should contain the exact command and environment needed to run to reproduce the results. See the NeurIPS code and data submission guidelines (\url{https://neurips.cc/public/guides/CodeSubmissionPolicy}) for more details.
        \item The authors should provide instructions on data access and preparation, including how to access the raw data, preprocessed data, intermediate data, and generated data, etc.
        \item The authors should provide scripts to reproduce all experimental results for the new proposed method and baselines. If only a subset of experiments are reproducible, they should state which ones are omitted from the script and why.
        \item At submission time, to preserve anonymity, the authors should release anonymized versions (if applicable).
        \item Providing as much information as possible in supplemental material (appended to the paper) is recommended, but including URLs to data and code is permitted.
    \end{itemize}

\item {\bf Experimental setting/details}
    \item[] Question: Does the paper specify all the training and test details (e.g., data splits, hyperparameters, how they were chosen, type of optimizer) necessary to understand the results?
    \item[] Answer: \answerNo{}
    \item[] Justification: Task inputs, score rules, and generation parameters are described; exact service-side defaults and all replay settings are not specified (Appendix~\ref{app:reproducibility}).
    \item[] Guidelines:
    \begin{itemize}
        \item The answer \answerNA{} means that the paper does not include experiments.
        \item The experimental setting should be presented in the core of the paper to a level of detail that is necessary to appreciate the results and make sense of them.
        \item The full details can be provided either with the code, in appendix, or as supplemental material.
    \end{itemize}

\item {\bf Experiment statistical significance}
    \item[] Question: Does the paper report error bars suitably and correctly defined or other appropriate information about the statistical significance of the experiments?
    \item[] Answer: \answerNo{}
    \item[] Justification: The reported aggregates do not include generation-repeat uncertainty or inferential error bars.
    \item[] Guidelines:
    \begin{itemize}
        \item The answer \answerNA{} means that the paper does not include experiments.
        \item The authors should answer \answerYes{} if the results are accompanied by error bars, confidence intervals, or statistical significance tests, at least for the experiments that support the main claims of the paper.
        \item The factors of variability that the error bars are capturing should be clearly stated (for example, train/test split, initialization, random drawing of some parameter, or overall run with given experimental conditions).
        \item The method for calculating the error bars should be explained (closed form formula, call to a library function, bootstrap, etc.)
        \item The assumptions made should be given (e.g., Normally distributed errors).
        \item It should be clear whether the error bar is the standard deviation or the standard error of the mean.
        \item It is OK to report 1-sigma error bars, but one should state it. The authors should preferably report a 2-sigma error bar than state that they have a 96\% CI, if the hypothesis of Normality of errors is not verified.
        \item For asymmetric distributions, the authors should be careful not to show in tables or figures symmetric error bars that would yield results that are out of range (e.g., negative error rates).
        \item If error bars are reported in tables or plots, the authors should explain in the text how they were calculated and reference the corresponding figures or tables in the text.
    \end{itemize}

\item {\bf Experiments compute resources}
    \item[] Question: For each experiment, does the paper provide sufficient information on the computer resources (type of compute workers, memory, time of execution) needed to reproduce the experiments?
    \item[] Answer: \answerNo{}
    \item[] Justification: API timing is reported, but the paper does not give complete hardware and memory specifications for every evaluation run.
    \item[] Guidelines:
    \begin{itemize}
        \item The answer \answerNA{} means that the paper does not include experiments.
        \item The paper should indicate the type of compute workers CPU or GPU, internal cluster, or cloud provider, including relevant memory and storage.
        \item The paper should provide the amount of compute required for each of the individual experimental runs as well as estimate the total compute. 
        \item The paper should disclose whether the full research project required more compute than the experiments reported in the paper (e.g., preliminary or failed experiments that didn't make it into the paper). 
    \end{itemize}
    
\item {\bf Code of ethics}
    \item[] Question: Does the research conducted in the paper conform, in every respect, with the NeurIPS Code of Ethics \url{https://neurips.cc/public/EthicsGuidelines}?
    \item[] Answer: \answerYes{}
    \item[] Justification: The study uses synthetic tasks and researcher curation; construction, execution risks, and asset terms are discussed in Section~\ref{sec:construction} and Appendices~\ref{app:impact} and~\ref{app:release}.
    \item[] Guidelines:
    \begin{itemize}
        \item The answer \answerNA{} means that the authors have not reviewed the NeurIPS Code of Ethics.
        \item If the authors answer \answerNo, they should explain the special circumstances that require a deviation from the Code of Ethics.
        \item The authors should make sure to preserve anonymity (e.g., if there is a special consideration due to laws or regulations in their jurisdiction).
    \end{itemize}

\item {\bf Broader impacts}
    \item[] Question: Does the paper discuss both potential positive societal impacts and negative societal impacts of the work performed?
    \item[] Answer: \answerYes{}
    \item[] Justification: Appendix~\ref{app:impact} discusses diagnostic benefits, misleading simulations, unsafe execution, and misuse.
    \item[] Guidelines:
    \begin{itemize}
        \item The answer \answerNA{} means that there is no societal impact of the work performed.
        \item If the authors answer \answerNA{} or \answerNo, they should explain why their work has no societal impact or why the paper does not address societal impact.
        \item Examples of negative societal impacts include potential malicious or unintended uses (e.g., disinformation, generating fake profiles, surveillance), fairness considerations (e.g., deployment of technologies that could make decisions that unfairly impact specific groups), privacy considerations, and security considerations.
        \item The conference expects that many papers will be foundational research and not tied to particular applications, let alone deployments. However, if there is a direct path to any negative applications, the authors should point it out. For example, it is legitimate to point out that an improvement in the quality of generative models could be used to generate Deepfakes for disinformation. On the other hand, it is not needed to point out that a generic algorithm for optimizing neural networks could enable people to train models that generate Deepfakes faster.
        \item The authors should consider possible harms that could arise when the technology is being used as intended and functioning correctly, harms that could arise when the technology is being used as intended but gives incorrect results, and harms following from (intentional or unintentional) misuse of the technology.
        \item If there are negative societal impacts, the authors could also discuss possible mitigation strategies (e.g., gated release of models, providing defenses in addition to attacks, mechanisms for monitoring misuse, mechanisms to monitor how a system learns from feedback over time, improving the efficiency and accessibility of ML).
    \end{itemize}
    
\item {\bf Safeguards}
    \item[] Question: Does the paper describe safeguards that have been put in place for responsible release of data or models that have a high risk for misuse (e.g., pre-trained language models, image generators, or scraped datasets)?
    \item[] Answer: \answerYes{}
    \item[] Justification: Offline request filtering and its limitations are described in Appendix~\ref{app:reproducibility}; it is not presented as host-isolation certification.
    \item[] Guidelines:
    \begin{itemize}
        \item The answer \answerNA{} means that the paper poses no such risks.
        \item Released models that have a high risk for misuse or dual-use should be released with necessary safeguards to allow for controlled use of the model, for example by requiring that users adhere to usage guidelines or restrictions to access the model or implementing safety filters. 
        \item Datasets that have been scraped from the Internet could pose safety risks. The authors should describe how they avoided releasing unsafe images.
        \item We recognize that providing effective safeguards is challenging, and many papers do not require this, but we encourage authors to take this into account and make a best faith effort.
    \end{itemize}

\item {\bf Licenses for existing assets}
    \item[] Question: Are the creators or original owners of assets (e.g., code, data, models), used in the paper, properly credited and are the license and terms of use explicitly mentioned and properly respected?
    \item[] Answer: \answerNo{}
    \item[] Justification: Source-specific terms and release differences are discussed; the paper does not provide a complete per-asset license ledger (Appendix~\ref{app:release}).
    \item[] Guidelines:
    \begin{itemize}
        \item The answer \answerNA{} means that the paper does not use existing assets.
        \item The authors should cite the original paper that produced the code package or dataset.
        \item The authors should state which version of the asset is used and, if possible, include a URL.
        \item The name of the license (e.g., CC-BY 4.0) should be included for each asset.
        \item For scraped data from a particular source (e.g., website), the copyright and terms of service of that source should be provided.
        \item If assets are released, the license, copyright information, and terms of use in the package should be provided. For popular datasets, \url{paperswithcode.com/datasets} has curated licenses for some datasets. Their licensing guide can help determine the license of a dataset.
        \item For existing datasets that are re-packaged, both the original license and the license of the derived asset (if it has changed) should be provided.
        \item If this information is not available online, the authors are encouraged to reach out to the asset's creators.
    \end{itemize}

\item {\bf New assets}
    \item[] Question: Are new assets introduced in the paper well documented and is the documentation provided alongside the assets?
    \item[] Answer: \answerNo{}
    \item[] Justification: Task documentation is public, but release snapshots differ in membership and license metadata (Appendix~\ref{app:release}).
    \item[] Guidelines:
    \begin{itemize}
        \item The answer \answerNA{} means that the paper does not release new assets.
        \item Researchers should communicate the details of the dataset\slash code\slash model as part of their submissions via structured templates. This includes details about training, license, limitations, etc. 
        \item The paper should discuss whether and how consent was obtained from people whose asset is used.
        \item At submission time, remember to anonymize your assets (if applicable). You can either create an anonymized URL or include an anonymized zip file.
    \end{itemize}

\item {\bf Crowdsourcing and research with human subjects}
    \item[] Question: For crowdsourcing experiments and research with human subjects, does the paper include the full text of instructions given to participants and screenshots, if applicable, as well as details about compensation (if any)?
    \item[] Answer: \answerNA{}
    \item[] Justification: The study involves researcher task curation, not crowdsourcing or human-subject experiments (Section~\ref{sec:construction}).
    \item[] Guidelines:
    \begin{itemize}
        \item The answer \answerNA{} means that the paper does not involve crowdsourcing nor research with human subjects.
        \item Including this information in the supplemental material is fine, but if the main contribution of the paper involves human subjects, then as much detail as possible should be included in the main paper. 
        \item According to the NeurIPS Code of Ethics, workers involved in data collection, curation, or other labor should be paid at least the minimum wage in the country of the data collector. 
    \end{itemize}

\item {\bf Institutional review board (IRB) approvals or equivalent for research with human subjects}
    \item[] Question: Does the paper describe potential risks incurred by study participants, whether such risks were disclosed to the subjects, and whether Institutional Review Board (IRB) approvals (or an equivalent approval/review based on the requirements of your country or institution) were obtained?
    \item[] Answer: \answerNA{}
    \item[] Justification: The study does not involve crowdsourcing or experiments with human subjects.
    \item[] Guidelines:
    \begin{itemize}
        \item The answer \answerNA{} means that the paper does not involve crowdsourcing nor research with human subjects.
        \item Depending on the country in which research is conducted, IRB approval (or equivalent) may be required for any human subjects research. If you obtained IRB approval, you should clearly state this in the paper. 
        \item We recognize that the procedures for this may vary significantly between institutions and locations, and we expect authors to adhere to the NeurIPS Code of Ethics and the guidelines for their institution. 
        \item For initial submissions, do not include any information that would break anonymity (if applicable), such as the institution conducting the review.
    \end{itemize}

\item {\bf Declaration of LLM usage}
    \item[] Question: Does the paper describe the usage of LLMs if it is an important, original, or non-standard component of the core methods in this research? Note that if the LLM is used only for writing, editing, or formatting purposes and does \emph{not} impact the core methodology, scientific rigor, or originality of the research, declaration is not required.
    \item[] Answer: \answerYes{}
    \item[] Justification: LLM-assisted expansion and model evaluation are described in Sections~\ref{sec:construction} and~\ref{sec:exp_setup}, rather than declared as editing only.
    \item[] Guidelines:
    \begin{itemize}
        \item The answer \answerNA{} means that the core method development in this research does not involve LLMs as any important, original, or non-standard components.
        \item Please refer to our LLM policy in the NeurIPS handbook for what should or should not be described.
    \end{itemize}

\end{enumerate}

\end{document}